\documentclass[conference]{IEEEtran}
\IEEEoverridecommandlockouts
\usepackage{cite}
\usepackage{amsmath,amssymb,amsfonts}

\usepackage{graphicx}
\usepackage{textcomp}
\usepackage{xcolor}
\usepackage{url}
\newcommand{\ie}{i.\,e.,\xspace}
\newcommand{\eg}{e.\,g.,\xspace}
\newcommand{\etal}{et~al.\xspace}

\usepackage[numbers]{natbib}
\usepackage[noend]{algpseudocode}
\usepackage{algorithm,algorithmicx}

\usepackage{etoolbox,xspace}
\algrenewcommand\algorithmicrequire{\textbf{Input:}}
\algrenewcommand\algorithmicensure{\textbf{Postcondition:}}

\newcommand{\MACone}{M_{\mathrm{AC1}}}
\newcommand{\MACtwo}{M_{\mathrm{AC2}}}

\usepackage{comment}
\usepackage{caption}
\usepackage{subcaption}
\usepackage{diagbox}

\usepackage{todonotes}

\def\BibTeX{{\rm B\kern-.05em{\sc i\kern-.025em b}\kern-.08em
    T\kern-.1667em\lower.7ex\hbox{E}\kern-.125emX}}
\begin{document}

\title{Lifelong Graph Learning for Graph Summarization}

\author{\IEEEauthorblockN{Jonatan Frank, Marcel Hoffmann,\\ Nicolas Lell}
\IEEEauthorblockA{\textit{Data Science and Big Data Analytics} \\
Ulm University, Germany \\
firstname.lastname@uni-ulm.de}
\and
%\IEEEauthorblockN{2\textsuperscript{st} }
%\IEEEauthorblockA{\textit{Data Science and Big Data Analytics} \\
%\textit{Ulm University}\\
%Ulm, Germany \\
%andor.diera@uni-ulm.de}
%\and
\IEEEauthorblockN{David Richerby}
\IEEEauthorblockA{
University of Essex, Colchester, UK \\
david.richerby@essex.ac.uk}
\and
\IEEEauthorblockN{Ansgar Scherp}
\IEEEauthorblockA{\textit{Data Science and Big Data Analytics} \\
Ulm University, Germany \\
ansgar.scherp@uni-ulm.de}
}

\maketitle
\begin{abstract}
Summarizing web graphs is challenging due to the heterogeneity of the modeled information and its changes over time. 
We investigate the use of neural networks for lifelong graph summarization. 
After observing the web graph at a point in time, we train a network to summarize graph vertices. 
We apply this trained network to summarize the vertices of the changed graph at the next point in time. 
Subsequently, we continue training and evaluating the network to perform lifelong graph summarization. 
We use the GNNs Graph-MLP and GCN, as well as an MLP baseline, to summarize the temporal graphs. 
We compare $1$-hop and $2$-hop summaries. 
We investigate the impact of reusing parameters from a previous snapshot by measuring backward and forward transfer as well as forgetting rate.
Our extensive experiments are on two series of ten weekly snapshots, from 2012 and 2022, of a web graph with over $100$M edges. They show that all networks predominantly use $1$-hop information to determine the summary, even when performing $2$-hop summarization. 
Due to the heterogeneity of web graphs, in some snapshots, the $2$-hop summary produces up to ten times as many vertex classes as the $1$-hop summary. 
When using the network trained on the last snapshot from 2012 and applying it to the first snapshot of 2022, we observe a strong drop in accuracy. 
We attribute this drop over the ten-year time warp to the strongly increased heterogeneity of the web graph in 2022.
The source code and additional resources are available at \url{https://github.com/jofranky/Lifelong-Graph-Summarization-with-Neural-Networks}.
\end{abstract}

\maketitle

\section{Introduction}

Graph summarization is the generation of a small representation $S$ of an input graph $G$, that preserves structural information necessary for a given task~\cite{DBLP:journals/vldb/CebiricGKKMTZ19}. 
Calculating the graph summary itself can be computationally expensive~\cite{DBLP:journals/tcs/BlumeRS21}. 
However, the reduction in size allows tasks such as web data summary search~\cite{DBLP:journals/ws/KonrathGSS12} and data visualization~\cite{DBLP:journals/vldb/GoasdoueGM20}, to be computed much faster than on the original graph~\cite{DBLP:journals/tcs/BlumeRS21}.
In machine learning, graph summarization is used to improve the scalability of node embedding learning methods~\cite{DBLP:conf/aaai/0002LKSC23}.

The information preserved by a summary is defined by the summary model and considers the vertices' $k$-hop neighborhoods.
Thus, in a machine learning sense, graph summarization can be expressed as a vertex classification task where each vertex in the graph belongs to a certain summary~\cite{DBLP:conf/dsaa/BlasiFHRS22}, representing the class.
This summarization operation has to be permutation invariant to the input, \ie the vertices' $k$-hop neighborhood~\cite{DBLP:journals/tcs/BlumeRS21,DBLP:journals/ws/KonrathGSS12}.
Graph Neural Networks (GNNs) are designed to be permutation invariant in this way~\cite{DBLP:conf/dsaa/BlasiFHRS22}. 

\citet{DBLP:conf/dsaa/BlasiFHRS22} applied multiple GNNs to graph summarization. 
They considered graph summaries based on vertex equivalence classes~\cite{DBLP:journals/vldb/GoasdoueGM20} which are lossless with respect to specific features, such as the edge label of the outgoing edges of a vertex. 
The authors used the 6~May, 2012 snapshot of the Dynamic Linked Data Observatory (DyLDO) dataset~\cite{DBLP:conf/www/KaferUHP12}.
We extend the analysis of \citet{DBLP:conf/dsaa/BlasiFHRS22} with a temporal component by considering ten consecutive snapshots of DyLDO from each of 2012 and 2022.
We evaluate reusing networks trained on 2012 for 2022, which we refer to as a time warp.

The problem with training a network only on the first snapshot is that performance decreases 
when the graph changes over time~\cite{DBLP:conf/aaai/0002C21}. 
This is especially true for graph summaries, where classes may appear or disappear between snapshots~\cite{DBLP:conf/dsaa/BlasiFHRS22}. 
One simple solution is learning the summaries for each snapshot from scratch, resulting in a separate network per snapshot without any information about summaries of previous snapshots.
In this work, we address this problem by incrementally learning the network.
We reuse the network trained on previous snapshots as a base for learning the summaries on the next snapshot.
This is called lifelong graph learning~\cite{DBLP:series/synthesis/2018Chen}.

We first calculate the summaries of the vertices of each snapshot using a crisp algorithm~\cite{DBLP:journals/tcs/BlumeRS21}.
These summaries are used as the gold standard for training the GNNs.
We continually train the networks on one snapshot after another.
After training on a snapshot, we evaluate the network on that snapshot, and all past and future snapshots.
We measure forward transfer, backward transfer, and forgetting rates of the networks based on classification accuracy~\cite{DBLP:conf/nips/Lopez-PazR17, DBLP:conf/eccv/ChaudhryDAT18}.
Forgetting is the performance drop of a network on snapshots trained on in earlier tasks after training on subsequent tasks~\cite{mccloskey1989catastrophic}.
Our results show that a network performs best on each snapshot when it is trained on the sequence of tasks up to and including that snapshot.
This observation is more prominent for the $2$-hop than the $1$-hop summary.
After the ten-year time warp, reusing the network parameters from 2012 neither improves nor harms the performance in 2022.
In summary, our contributions are:
\begin{itemize}
% To time
    \item We extend graph summarization using GNNs from static  
    to temporal graphs. 
    We experiment with two GNNs and a baseline MLP on sequences of ten weekly snapshots from 2012 and 2022 of the DyLDO web graph. 

% Lifelong training
    \item We show that neural networks perform best on each snapshot when being trained on a sequence of tasks up to and including that snapshot.
    Changes in the graph reduce the performance which is more prominent in 2022.

%2-hop information does not help
    \item  
    We show that an MLP using only $1$-hop information is sufficient for $2$-hop summaries.
    
% The time warp
    \item 
    %on the last snapshot 
    Reusing parameters from a network trained in 2012 has no benefit over a network trained from scratch  
    %the on first snapshot 
    in 2022.
       
\end{itemize}

The remainder of the paper is organized as follows: 
Below, we summarize related work.
Section~\ref{sec:methods} introduces graph summarization. 
Section~\ref{sec:experimentalapparatus}  describes the experimental apparatus.
Our results are presented in Section~\ref{sec:results}. 
Section~\ref{sec:discussion} discusses the results, before we conclude.

\section{Related Work}
\label{sec:relatedwork}
\subsection{Graph Neural Networks}\label{subsec:GNN}

GNNs use the structural information of a graph contained in its edges and the features of the neighboring vertices to distinguish different vertices, \eg to classify them.
Many GNNs, including graph convolutional networks (GCNs)~\cite{DBLP:conf/iclr/KipfW17}, graph attention networks (GANs)~\cite{DBLP:conf/iclr/VelickovicCCRLB18} and GraphSAINT~\cite{DBLP:conf/iclr/ZengZSKP20} use message passing. 
GraphSAINT is a scalable GNN architecture that samples smaller subgraphs to enable any GNN to be trained on large graphs.
\citet{DBLP:journals/corr/abs-2106-04051} introduced Graph-MLP, a GNN that does not use message passing.
Graph-MLP consists of a multi-layer perceptron (MLP) that is trained with a cross-entropy loss and a neighbor contrastive (NContrast) loss on the graph edges. 

\subsection{Graph Summarization}
\label{subsec:summarization}
FLUID~\cite{DBLP:journals/tcs/BlumeRS21} is a language and a generic algorithm for flexibly defining and efficiently computing graph summaries. 
It can express all existing lossless structural graph summaries w.r.t. the considered features, such as the vertices’ edges. %, the vertex neighbors, and others. 
Computation of graph summaries is based on hash functions applied on a canonical order of the vertices features~\cite{DBLP:journals/tcs/BlumeRS21,DBLP:journals/ws/KonrathGSS12}. 
The use of neural networks as a hash function has mainly been in the context of security~\cite{7747793, DBLP:journals/corr/abs-0707-4032}, but recently \citet{DBLP:conf/dsaa/BlasiFHRS22} applied GCNs, GraphSAGE~\cite{DBLP:conf/nips/HamiltonYL17}, GraphSAINT and Graph-MLP to summarizing static graphs, along with Blooom filters.
We use a GraphSAINT-based sampling method by \citet{DBLP:conf/dsaa/BlasiFHRS22}, which considers the class distribution by sampling inverse to the number of occurrences of classes in the training set.

\subsection{Lifelong Graph Learning}\label{subsec:LlGL}
Lifelong learning \cite{DBLP:series/synthesis/2018Chen}
is a learning procedure that imitates the lifelong learning ability of humans.  
It adds the importance of transferring and refining knowledge to the learned network. 
In lifelong learning~\cite{DBLP:conf/kdd/FeiW016}, a network at time $t$ has performed a sequence of $t$ learning tasks $\mathcal{T}_1, \mathcal{T}_2, \dots, \mathcal{T}_t$ and has accumulated knowledge from these past tasks. 
At time $t+1$, it is faced with a new learning task $\mathcal{T}_{t+1}$. 
The network can use past knowledge to help with learning task~$\mathcal{T}_{t+1}$.

A lifelong learning model should be able to exploit previous knowledge to learn new tasks better or faster.
One issue in lifelong learning is catastrophic forgetting~\cite{mccloskey1989catastrophic}. 
Catastrophic forgetting is the tendency of neural networks to drastically forget previous knowledge upon learning new information. 

In lifelong graph learning, a challenge is that the graphs can grow or shrink over time.
Adding and removing vertices in the graphs may also result in the inclusion of new class labels in classification tasks~\cite{DBLP:conf/aaai/0002C21}. 
We address the aspect of increasing the detection accuracy of already-seen classes and adding new classes to our networks.
\citet{galke-neuralnetworks-journal} 
compare reusing an existing network (warm restart) versus training a new network from scratch (cold restart).
Their experiments show that warm restarts are generally preferred for lifelong graph learning.

\section{Graphs and Summarization}
\label{sec:methods}
We define the basic notation of graphs used as input for our summary models. 
We introduce classical graph summarization, which is used to compute the gold standard, and graph summarization with neural networks.
We describe our sampling approach and our training method. 
Finally, we describe how graph summarization is used in the lifelong learning setting.

\subsection{Labeled Graphs}

We consider web graphs with labeled relations of the form $G = (V, E, R)$ consisting of a set of vertices $V$, a set of relation types $R$, and a set of labeled edges $E \subseteq V \times R \times V$ that express relationships between entities in $V$. 
This allows graphs to be represented in the Resource Description Framework (RDF),\footnote{\url{https://www.w3.org/RDF}} a W3C standard that models relations in the form of subject--predicate--object triples $(s,p,o)$~\cite{rdf1}. 
\texttt{rdf:type} is a special predicate in the RDF standard. 
It assigns a vertex $s$ a label $o$, \ie a vertex type. 
This is done by the triple $(s, \texttt{rdf:type}, o)$. 
The predicates $p\neq \texttt{rdf:type}$ are called RDF properties. 
Our summary models focus on the properties and do not consider \texttt{rdf:type} following \citet{DBLP:conf/esws/GottronKSS13}.
They argue that $20\%$ of all data providers of web graphs do not need RDF types because the vertices are already precisely described by the edges.

\subsection{Classical Graph Summarization}

A graph summary maps a graph $G$ to a smaller representation $S$, which is a graph that preserves structural information of $G$~\cite{DBLP:journals/tgdk/ScherpRBCR23}.
Each vertex of the original graph $G$ is a member of exactly one equivalence class (EQC) of the summary graph. Each EQC represents a set of vertices that share common features, such as the same set of edge labels. 
We formulate graph summarization as a vertex classification task following \citet{DBLP:conf/dsaa/BlasiFHRS22}. 
Our summary models depend on the $k$-hop neighborhood of a vertex.
We use the $1$- and $2$-hop versions of Attribute Collection, $\MACone$ and $\MACtwo$, respectively. $\MACone$ considers a vertex's RDF properties and $\MACtwo$ also includes the neighbors' properties.
These models use of outgoing edges and only differ in the size of the neighborhood that is considered for the summary. 
In $M_{\mathrm{AC}k}$, we calculate the EQC of a vertex~$x$ by calculating the following function $h_k(x)$ that considers the $k$-hop neighborhood of the vertex.
We set $h_1(x)=0$ and then, for $k\geq 1$,
\begin{equation*}
   \label{eq:hashing}
   h_{k+1}(x)\ \; =\!\! \bigoplus_{(x,p,y)\in E}\!\! \mathrm{hash}(p, h_k(y))\,,
\end{equation*}
where $\bigoplus$  is the bitwise XOR operator and the function $\mathrm{hash}(p, h_i(y))$ first concatenates $p$ and $h_i(y)$ to a string, which is then hashed.
We use the default hashing function for strings in Python, which uses SipHash~\cite{hash, aumasson2012siphash}.

\begin{figure}[ht]
\centering
\includegraphics[width=0.3\textwidth]{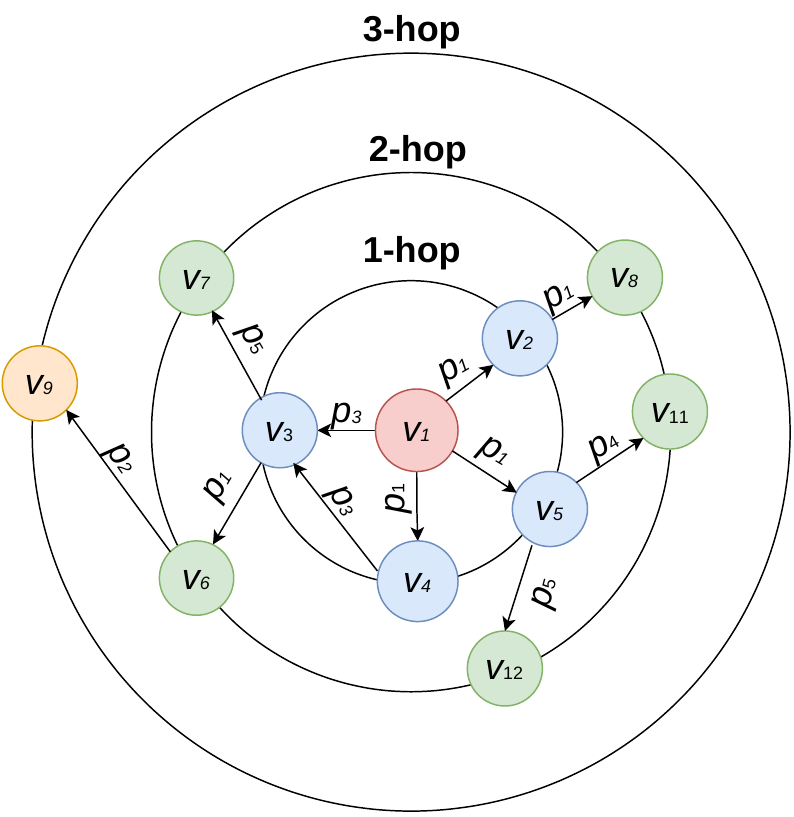}
\caption{$3$-hop neighborhood of a vertex $v_1$.}\label{fig:nhop}
\end{figure}

Figure~\ref{fig:nhop} illustrates summarizing a vertex.
Our model considers, at most, the $2$-hop neighborhood of a vertex. 
In the case of vertex $v_1$ (center), we recursively consider edges in the $2$-hop neighborhood of vertex $v_1$. Hence, we consider each edge in Figure \ref{fig:nhop} except the edge $(v_6, p_2, v_9)$, which extends to the three-hop neighborhood.

We generalize this classical graph summarization by considering graphs that change over time.
For a number of hops $i\in\{1,2\}$, the graph summary $S_t$ of a graph $G_t=(V_t,E_t,R_t)$ at time~$t$ is a tuple $(V^S_t,E^S_t,R^S_t)$. 
$V^S_t = C \cup D$, 
where $C$ is the set of EQCs computed from the vertices $V_t$, and $D=\{d_{p,h_i(y)} \mid (x,p,y)\in E_t \text{ for some }x\}$ is a set of vertices further specifying the EQCs. 
These are the primary and secondary vertices of Blume \etal~\cite{DBLP:journals/tcs/BlumeRS21}.
$R^S_t$ is the set of predicates of the original graph $G_t$. 
For $d_{p,h_i(y)}\in D$, the outgoing edges $E^S_t$ of the summary vertices $V^S_t$ are given by
%\[
$E^S_t = \{(v, p, d_{p,h_i(y)}) \mid (x,p,y)\in E_t \text{ and } x\in C\}$.
%\]

\subsection{Neural Graph Summarization}

When applying graph neural networks (GNNs) for graph summarization, we need to define the vertex features~\cite{DBLP:series/synthesis/2020Hamilton,DBLP:conf/dsaa/BlasiFHRS22}.
Instead of hashing, as above, the feature vector $h_v^{(0)}$ of a vertex~$v$ is a multi-hot encoding of its outgoing edge labels, following \citet{DBLP:conf/dsaa/BlasiFHRS22}.
The motivation is that the multi-hot encoded feature vector can be easily extended with new elements once new predicates appear in the temporal graph.
We note that multi-hot encoding is appropriate for graph summarization, as graph summary models only depend on the existence of edge or vertex labels, not their multiplicity~\cite{DBLP:journals/tcs/BlumeRS21}.

For the $1$-hop model $\MACone$, the feature vector already contains all the information needed to calculate the summary, which is the label set of the vertex's edges.
For the $2$-hop model $\MACtwo$, we consider the feature vector of $v_1$ and, in addition, consider the feature vectors of $v_1\!$'s neighbors.
From these feature vectors, we know the labels of the outgoing edges of $v_1$ and its neighbors.
Note that, in classical summarization, we also know which edge labels connect to which neighboring feature vectors.
Here, we know the set of edge labels and the set of neighboring feature vectors but not which edge is associated with which neighbor. 
%However, the neural networks can still differentiate between the equivalence classes in $\MACtwo$.

An approach to incorporate the edges is to represent an edge between a vertex and its neighbor as a vertex between them, labeled with the edge label one-hot encoded as a feature vector. 
In this case, we use two hidden layers in the GNN to cover the additional hop. 
Where this procedure is not applied, we use one hidden layer. 
We evaluate both approaches, i.\,e., test whether message passing is necessary for a $2$-hop summary

For lifelong learning, we consider different snapshots of a graph $G_t$ associated with a timestamp~$t$.
At time $t$, we train a neural network to predict the summary $S_{t}$ of the graph $G_t$ at time~$t$, which we refer to as task $\mathcal{T}_{t}$.
We compute the number of EQCs present in the summary of time~$t$ for each time~$t$, and the number of added and deleted EQCs compared to time~$t-1$.
From this, we can also compute how many EQCs from time $t-1$ are still present at time~$t$.
We also count how many different EQCs in total have appeared in the snapshots $S_1,\dots, S_{t}$, which corresponds to the size of the neural network's output layer at time~$t$.

We use two GNNs in our study: 
a GCN~\cite{DBLP:conf/iclr/KipfW17}, a classical message-passing model, and Graph-MLP~\cite{DBLP:journals/corr/abs-2106-04051}, which is an alternative to the usual message-passing GNNs.

Due to the large size of our graphs, we use the sampler of \citet{DBLP:conf/dsaa/BlasiFHRS22} for training.  
The sampler is based on the GraphSAINT vertex sampler, with each vertex weighted according to the class distribution.
This takes into account the imbalanced class distributions of the snapshots.

\subsection{Complexity Analysis}

Our approach has two steps: first, summarizing a graph w.r.t.\@ a given summary model and, second, training the graph neural network. 
For the first step, we iteratively compute summaries of a snapshot-based temporal graph.
This can be computed with an incremental, parallel algorithm in time $\mathcal{O}(N\,\Delta^k)$~\cite{DBLP:conf/cikm/BlumeRS20}, where $N$ is the number of changes in the graph from one snapshot to the next, $\Delta$~is the maximum degree, and $k$~is the number of hops considered in the summary model -- for us, $k\in\{1,2\}$.
This applies to summary models definable in FLUID, excluding any optional preprocessing step involving inference.
The approach is parallelized and scales to large graphs.

The complexity of the second step depends on the specific neural network, the sample of vertices $V' \subseteq V$, number of layers~$K$, and hidden dimension~$d$. 
Training and inference time is  
$\mathcal{O} (Kd(|E| + |V|d))$ for GCN~\cite{DBLP:conf/iclr/KipfW17}, 
training time is $\mathcal{O} (Kd(|E| + |V|d))$ and inference time is $\mathcal{O}(Kd)$ for Graph-MLP~\cite{DBLP:journals/corr/abs-2106-04051}, and 
$\mathcal{O}(Kd)$ for MLP~\cite{DBLP:journals/corr/abs-2106-04051}.

\section{Experimental Apparatus}
\label{sec:experimentalapparatus}
In this section, we introduce and analyze our dataset, and describe our experimental procedure. 
Afterward, we present how we optimize our hyperparameters. 
Finally, we present performance and forgetting measures for the evaluation of lifelong graph learning.

\subsection{Datasets: Weekly DyLDO Snapshots from 2012 and 2022}
\label{sec:datasets}

We use the weekly snapshots of the DyLDO web crawl as foundation~\cite{DBLP:conf/www/KaferUHP12}.
Starting from a seed of $90,000$ Uniform Resource Identifiers, it samples data from the web to create a snapshot. 
A snapshot consists line-based, plain text encoding of an Resource Description Framework (RDF) graph~\cite{rdf2}.

We create two datasets, using the following sequences of snapshots:
The ten snapshots from 6~May, 2012 to 8~July, 2012 (the first ten crawls) and the ten snapshots from 25~September, 2022 to 27~November, 2022.
Figures~\ref{AllEQCs2012m} and~\ref{AllEQCs2022m} show the number of EQCs in each snapshot.

For each snapshot, we create the labels of the vertices by calculating their EQCs according to the summary model.
For $\MACtwo$ we remove all vertices with a degree over $100$ from the training, validation, and test sets to speed up the experiments.

This has only a small effect on the number of EQCs: for example, in the first snapshot of DyLDO in 2012 only $12,425$ out of $288,418$ EQCs are removed, which is less than~$5\%$.

\subsection{Dataset Analysis}
\begin{figure}
     \centering
       \begin{subfigure}[b]{0.24\textwidth}
         \centering
         \includegraphics[width=\textwidth]{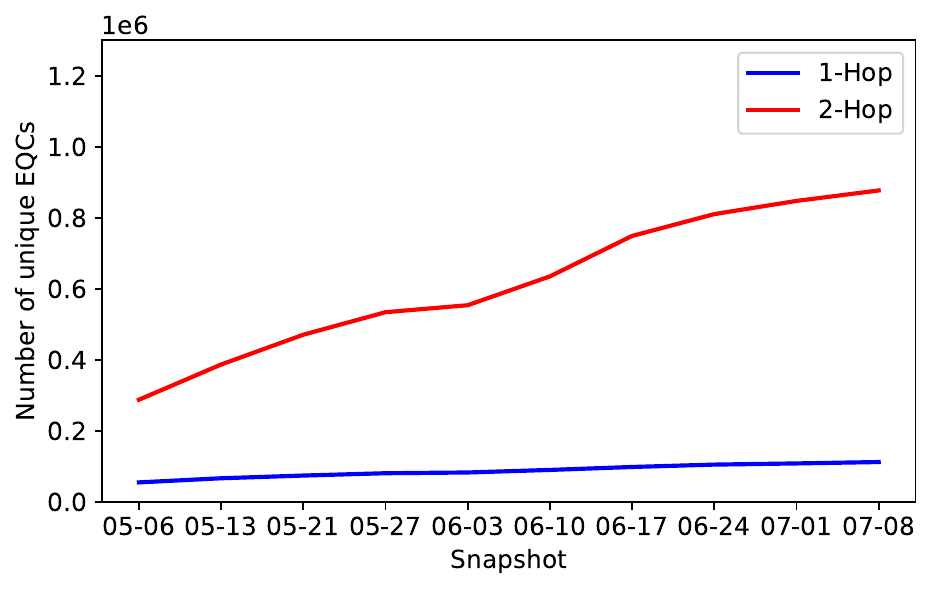}
          \caption{May--July 2012}
         \label{AllEQCs2012m}
     \end{subfigure}  \hfill
     \begin{subfigure}[b]{0.24\textwidth}
         \centering
         \includegraphics[width=\textwidth]{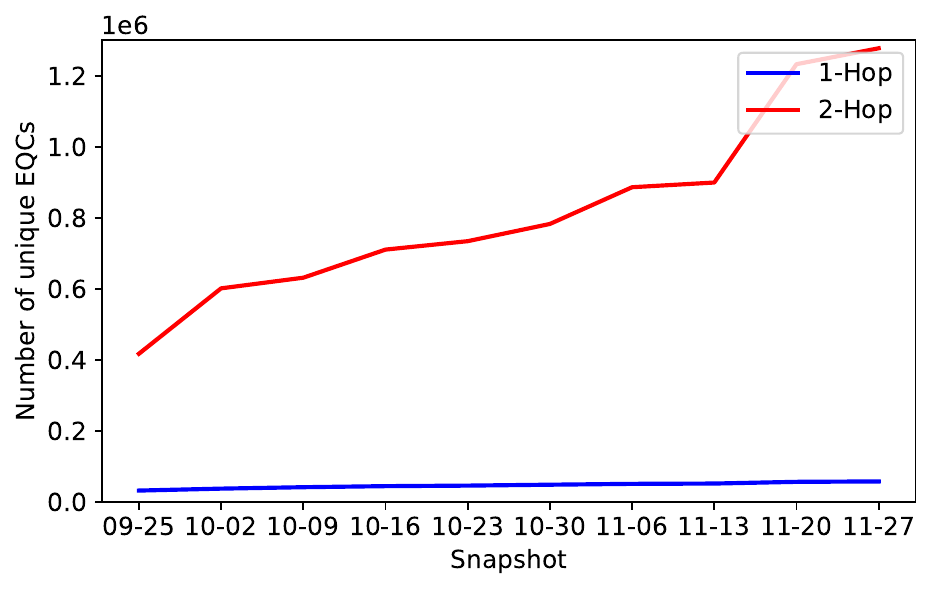}
          \caption{ September--November 2022}
         \label{AllEQCs2022m}    
     \end{subfigure}  \hfill
     \begin{subfigure}[b]{0.24\textwidth}
         \centering
         \includegraphics[width=\textwidth]{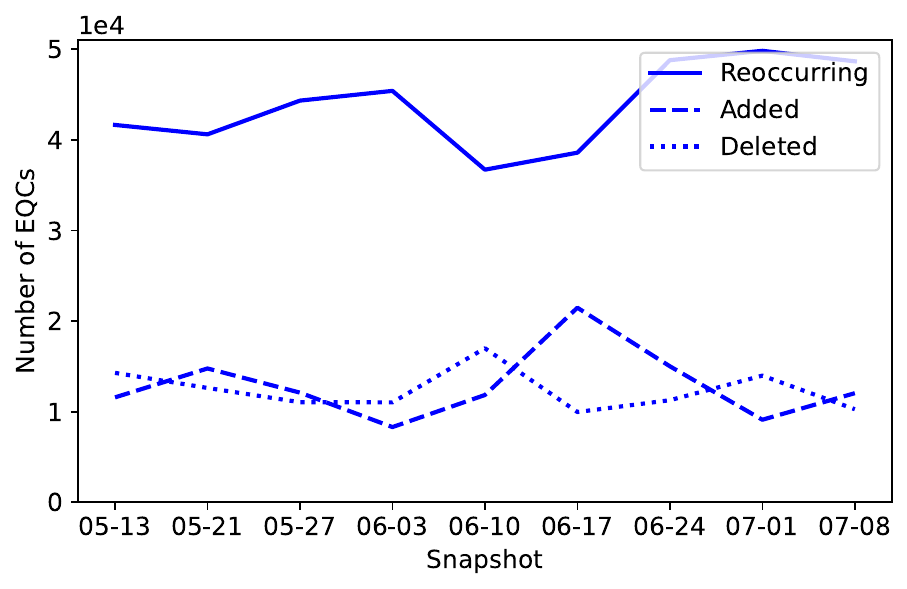}
         \caption{$1$-hop, May--July 2012}
         \label{ChangesP2012-1m}
     \end{subfigure}  \hfill
     \begin{subfigure}[b]{0.24\textwidth}
         \centering
         \includegraphics[width=\textwidth]{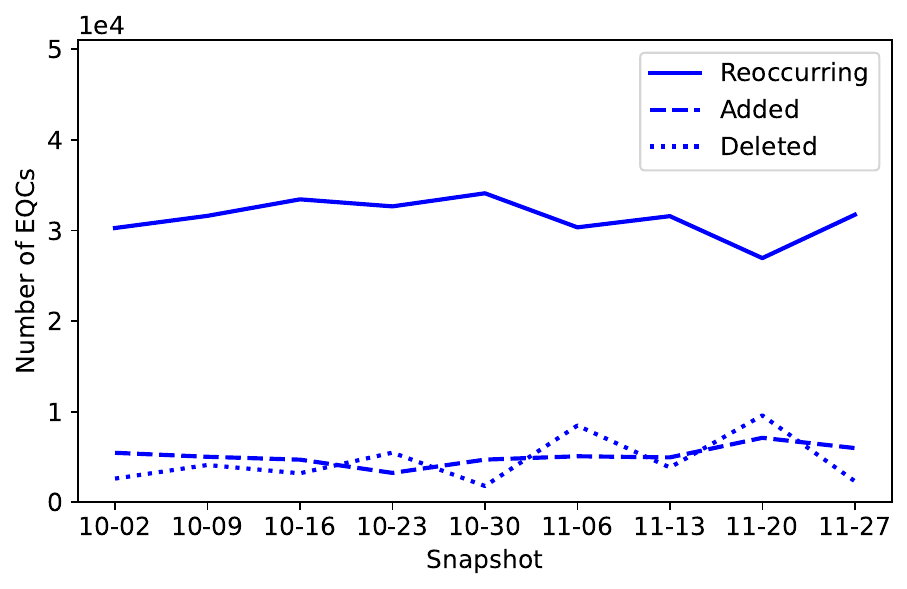}
         \caption{$1$-hop, Sept.--Nov.~2022}
         \label{ChangesP2022-1m}
     \end{subfigure}  \hfill
     \begin{subfigure}[b]{0.24\textwidth}
         \centering
         \includegraphics[width=\textwidth]{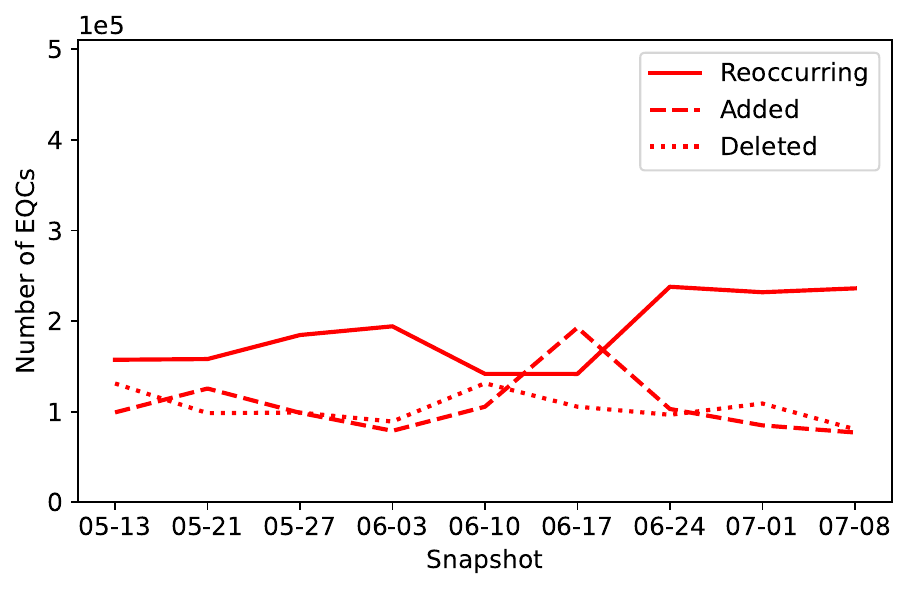}
         \caption{$2$-hop, May--July 2012}
         \label{ChangesP2012-2m}
     \end{subfigure}  \hfill
     \begin{subfigure}[b]{0.24\textwidth}
         \centering
         \includegraphics[width=\textwidth]{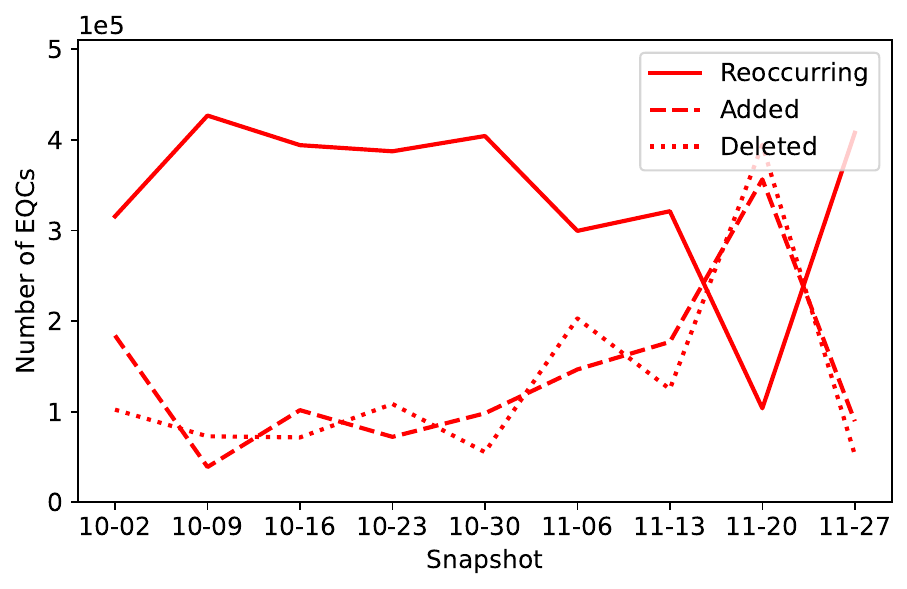}
         \caption{$2$-hop, Sept.--Nov.~2022}
         \label{ChangesP2022-2m}
     \end{subfigure}  \hfill
        \caption{Detailed analyses of the DyLDO snapshots used in our experiments.
        Figures~(a) and (b) show the number of unique EQCs per snapshot for a $1$-hop versus $2$-hop model.
        Figures~(c)--(f) show the changes of the EQCs (addition, deletion, and recurring)  compared to the previous snapshot.}
        \label{fig:dataA}
\end{figure}

Figure~\ref{fig:dataA} shows how the EQCs in each year change from snapshot to snapshot.
We observe that the EQCs of the $2$-hop summary change more than those of the $1$-hop summary. 
Both changes underlie the same snapshot modification, indicating that the $2$-hop summary changes more than the $1$-hop summary in the same sequence of snapshots.
In general, there are more EQCs in the $2$-hop summaries in 2022 than in 2012. 
The numbers of EQCs in the $1$-hop summaries are similar between the two years.

\subsection{Procedure}
\label{sec:procedure}
\paragraph{Training of GNNs and Baselines}
Based on pre-experiments, we train all networks with $100$ batches, each containing up to $1,000$ vertices.
Our baseline is a standard MLP with one hidden layer and ReLU-activation functions with dropout. 
For $\MACone$, we use an MLP because all information is contained in the feature vector. 
We have two cases for $\MACtwo$, one with encoded edge information and one without.  
In the first case, we only use GCN because the other neural networks do not use neighborhood information during inference.
In the second case, we use all neural networks, \ie GCN, MLP, and Graph-MLP.

\paragraph{Lifelong Graph Learning for Summarizing the DyLDO Snapshots}
We perform the lifelong graph learning experiments once for the $2012$ and once for the $2022$ snapshots.
We train the networks on the snapshots in chronological order.
We split the vertices of each snapshot into $2\%$ validation, $5\%$ testing and sample the training data from the remaining $93\%$.

For the first snapshot of each year, the neural networks are initialized with an input size of number of predicates and output size of the number of classes.
Then, the networks are trained on the snapshot for $100$ iterations with maximum batch size $1,000$.
After training, we increase the input and output size if the next snapshot has more classes (EQCs) or predicates.
The extended neural network is trained on the next snapshot for $100$ iterations. 
We repeat this process until the last snapshot. 
After training the networks on a snapshot, we evaluate them on all snapshots of the same year.
We do this to calculate the lifelong learning measures (see Section~\ref{sec:measures}). 

\paragraph{Time Warp Experiment}
In this experiment, we compare three networks on the first snapshot of 2022.
One ``10-year-old'' network is only trained on the 2012 snapshots, one is initialized from the old network but also trained on the 2022 snapshot, and one which is trained from scratch just on the 2022 snapshot.
This experiment shows whether reusing old networks of 2012 is helpful despite the time gap of ten years, compared to starting training from scratch in 2022.

\subsection{Hyperparameter Optimization}
\label{sec:hyperparameteroptimization}

For the hyperparameter search, we apply grid search.
We tune the learning rate on the values  $\{0.1, 0.01, 0.001\}$, dropout in $\{0.0, 0.2, 0.5\}$, and  hidden layer size $\{32, 64\}$ for the GNN. 
For Graph-MLP, we optimize the weighting coefficient $\alpha \in\{1.0, 10.0, 100.0\}$, the temperature $\tau \in  \{0.5, 1.0, 2.0\}$, learning rate in $\{0.1, 0.01\}$, and the hidden layer size $\{64, 256\}$. 
For MLP, we use a hidden layer size of $1,024$ following prior works~\cite{DBLP:journals/corr/abs-2109-03777, DBLP:conf/dsaa/BlasiFHRS22}.
We use Adam as an optimizer.

The best configuration depends on the summary model. 
Details can be found in our appendix.
For the summary model $\MACone$, we use an MLP with a hidden size of $1,024$, a dropout of $0.5$, and a learning rate of $0.01$.
For $\MACtwo$ with edge information, we use GCN with two hidden layers of size $32$, learning rate~$0.1$, and no dropout.
For $\MACtwo$ with no edge information, we use an MLP with one hidden layer of size $1,024$, dropout~$0.5$, and learning rate~$0.01$.
We use Graph-MLP with hidden size~$64$, dropout~$0.2$, learning rate~$0.01$, $\alpha = 1$, and $\tau = 2$.
GCN uses one hidden layer of dimension $64$, learning rate~$0.1$, and no dropout.

\subsection{Measures}
\label{sec:measures}

We assess vertex classification performance using test accuracy,  a well-known measure in graph neural networks like GCN~\cite{DBLP:conf/iclr/KipfW17}.
For lifelong learning, we use the measures from \citet{DBLP:conf/nips/Lopez-PazR17} and \citet{DBLP:conf/eccv/ChaudhryDAT18}.

We train a network for each snapshot and test it on all other snapshots.
Thus, for each of the years 2012 and 2022, we compute a result matrix $R \in \mathbb{R}^{T \times T}$ with $T=10$.
Each entry $R_{i,j}$ is the test accuracy of the network on task $\mathcal{T}_j$ after observing the last sample from task $\mathcal{T}_i$. 
That means the network is trained on tasks $\mathcal{T}_1$ to $\mathcal{T}_i$ and tested on $\mathcal{T}_j$.
We use these matrices to calculate the following measures for each year, to measure the backward and forward transfer of the networks.
The average accuracy (ACC) a network achieves on all tasks $\mathcal{T}_i$ after being trained on the last task $\mathcal{T}_T$ is given by $ACC = \frac{1}{T}\sum^T_{i=1}R_{T,i}$.
The backward transfer (BWT), which indicates how much  knowledge of previous tasks the network keeps after being trained on the last task $T$ is given by $BWT = \frac{1}{T-1}\sum^{T-1}_{i=1}(R_{T,i}-R_{i,i})$.
The forward transfer (FWT), which shows how much knowledge of a task is reused for the next one, is based on \citet{DBLP:conf/nips/Lopez-PazR17} as $FWT= \frac{1}{T-1} \sum_{i=2}^T (R_{i-1,i}-R_{i,i})$.

We define the ideal performance as $\alpha_\mathrm{ideal} = \max_{i \in \{1,\dots, T\}} R_{i,i}$.
We define $\alpha_{i,\mathrm{all}} = \frac{1}{T}\sum^{T}_{t=1} R_{i,t}$ as the average accuracy of a network trained up to task~$\mathcal{T}_i$.
With this, we can define the following measures, as in~\cite{DBLP:conf/aaai/KemkerMAHK18}:
$    \Omega_\mathrm{base} = \frac{1}{T-1}\sum^T_{i=2}\frac{R_{i,1}}{\alpha_\mathrm{ideal}}\text{, }
    \Omega_\mathrm{new} = \frac{1}{T-1}\sum^T_{i=2}R_{i,i}\text{, and }\Omega_\mathrm{all} = \frac{1}{T-1}\sum^T_{i=2}\frac{\alpha_{i,\mathrm{all}}}{\alpha_\mathrm{ideal}} \,.$  
The first snapshot is the base test set, and the best accuracy $\alpha_\mathrm{ideal}$ in the matrix is the optimum.
$\Omega_\mathrm{base}$ measures how a network performs on the base test set after training on other sets compared to the optimum. $\Omega_\mathrm{new}$ shows how a network performs on a snapshot after being trained on it. 
$\Omega_\mathrm{all}$ measures how a network performs on all previous snapshots compared to the optimum.

Finally, we compute the forgetting measure~\cite{DBLP:conf/eccv/ChaudhryDAT18}: $F_k = \frac{1}{k-1}\sum^{k-1}_{j=1}f_j^k$,
where $f_j^k$ is the forgetting on task $j$ after the network is trained up to task $k$ and is computed as:
$f_j^k = \max_{\ell \in \{1,\dots,k-1\}}(R_{\ell,j}-R_{k,j})$.

\section{Results}
\label{sec:results}

We describe the results for the vertex classification and the lifelong learning measures. 
Finally, we show the results of the ten-year time warp.

\paragraph{Vertex Classification}

 \begin{figure}
     \centering
     \begin{subfigure}[b]{0.375\textwidth}
         \centering
         \includegraphics[width=\textwidth]{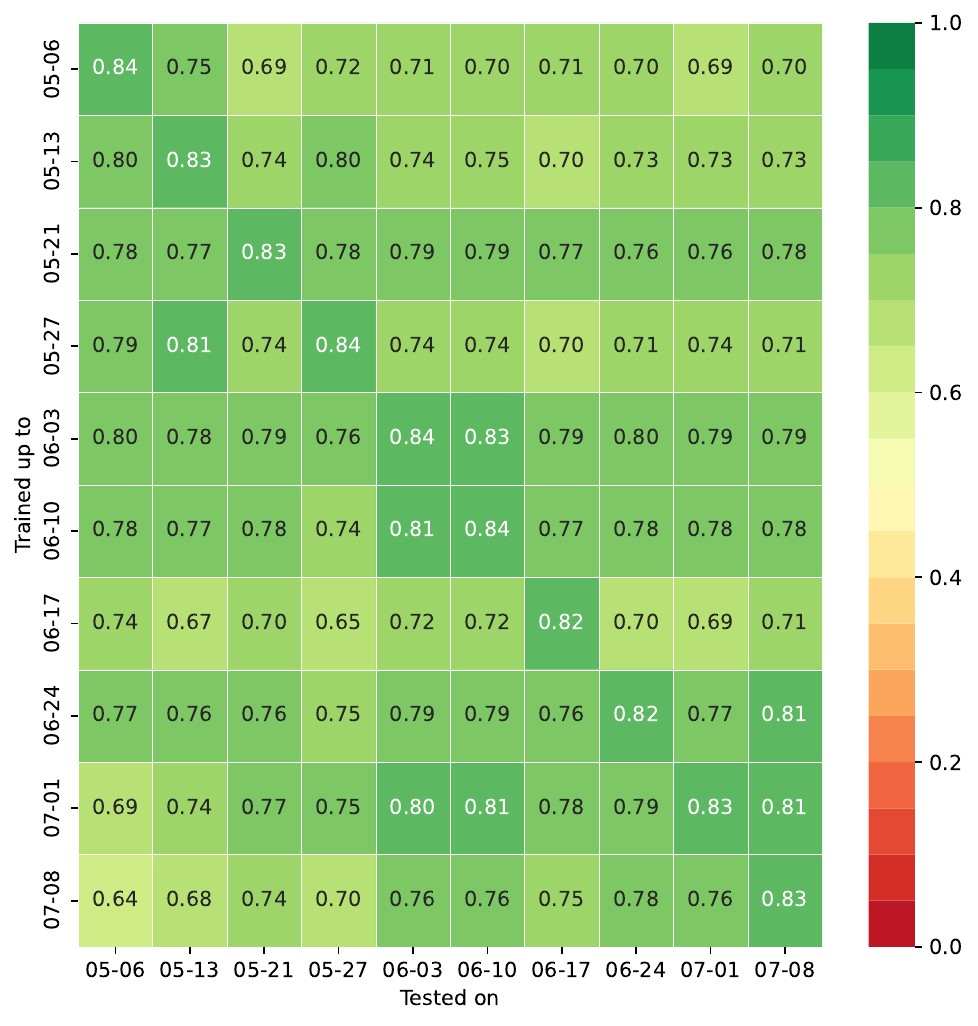}
         \caption{ MLP ($1$-hop)}
         \label{HeatmapMLP1_2012}
     \end{subfigure} 
     \begin{subfigure}[b]{0.375\textwidth}
         \centering
         \includegraphics[width=\textwidth]{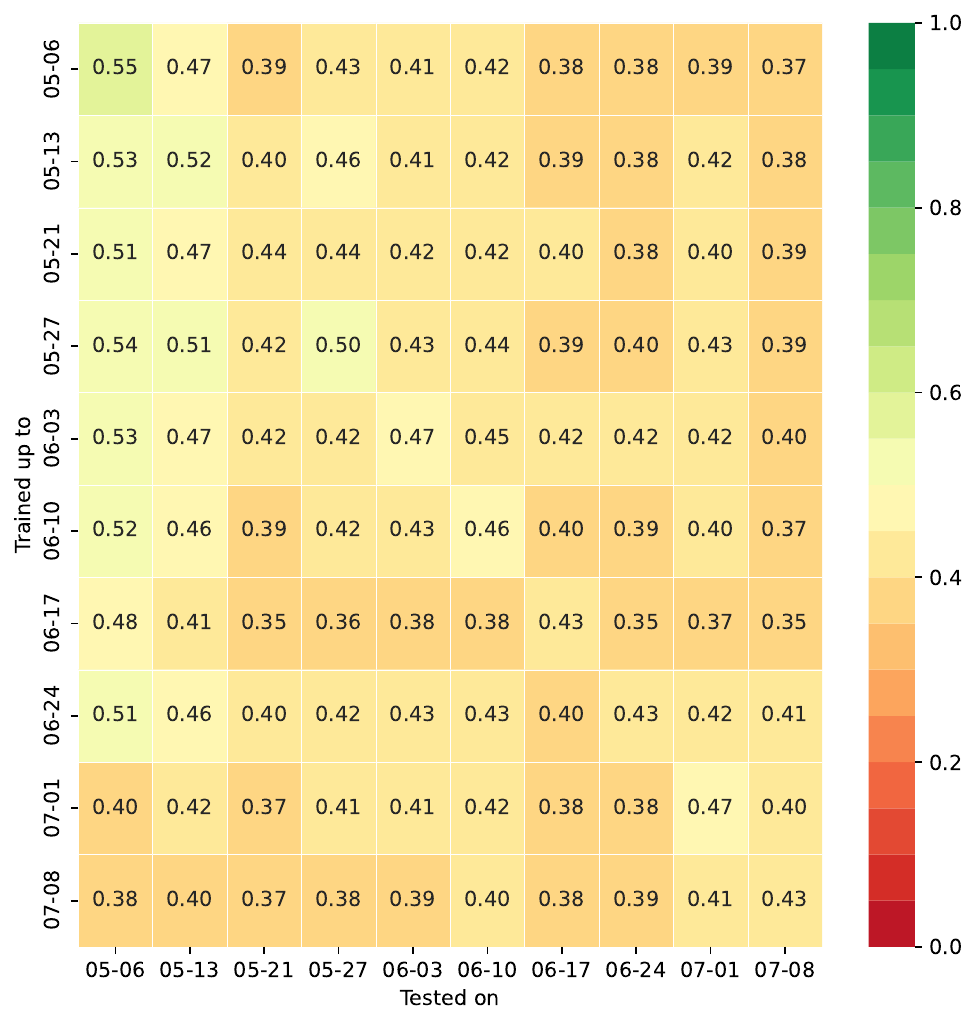}
         \caption{ MLP ($2$-hop)}
         \label{HeatmapMLP2_2012}
     \end{subfigure} 
        \caption{Accuracies for snapshots trained from May to July 2012. The accuracies for Graph-MLP ($2$-hop), GCN ($2$-hop), and GCN ($2$-hop and edges) are omitted because they are similar to the values of the  MLP ($2$-hop). }
        \label{fig:heatmaps2012}
\end{figure}

 \begin{figure}
     \centering
     \begin{subfigure}[b]{0.375\textwidth}
         \centering
         \includegraphics[width=\textwidth]{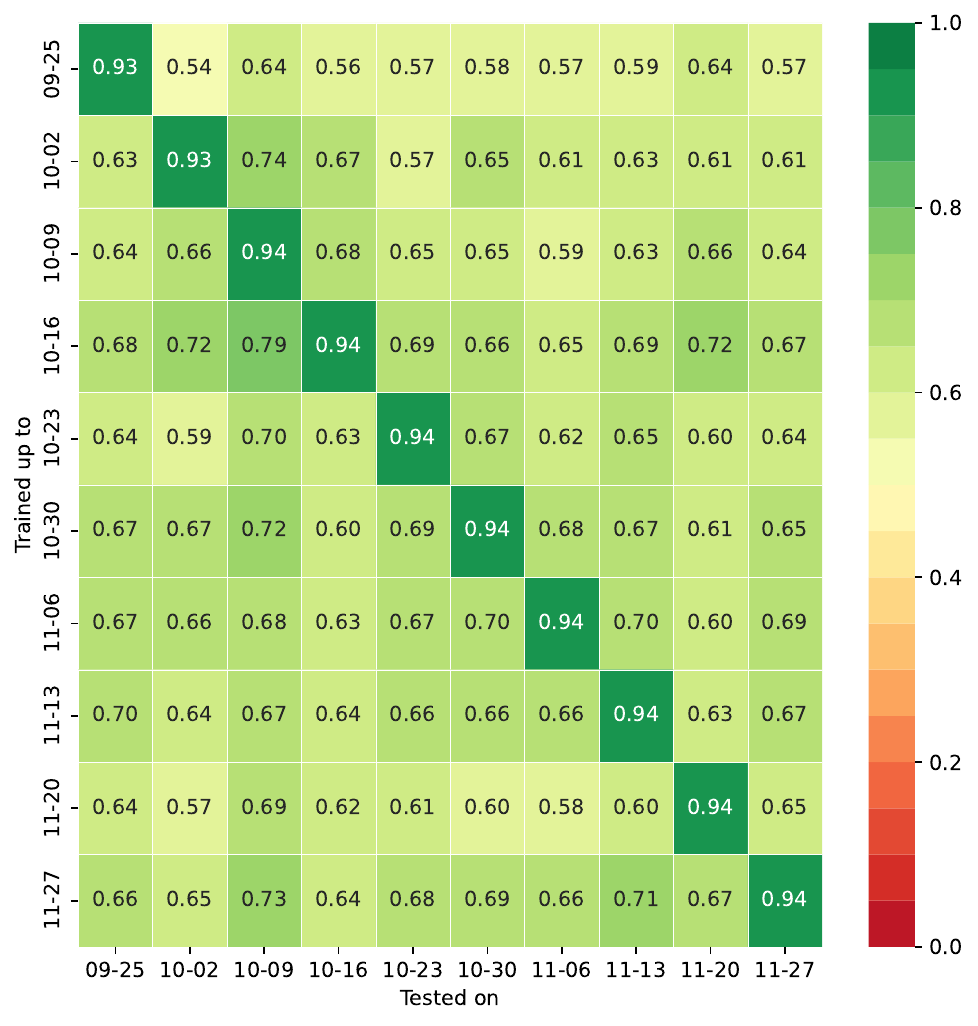}
         \caption{  MLP ($1$-hop)}
         \label{HeatmapMLP1_2022}
     \end{subfigure}  
     \begin{subfigure}[b]{0.375\textwidth}
         \centering
         \includegraphics[width=\textwidth]{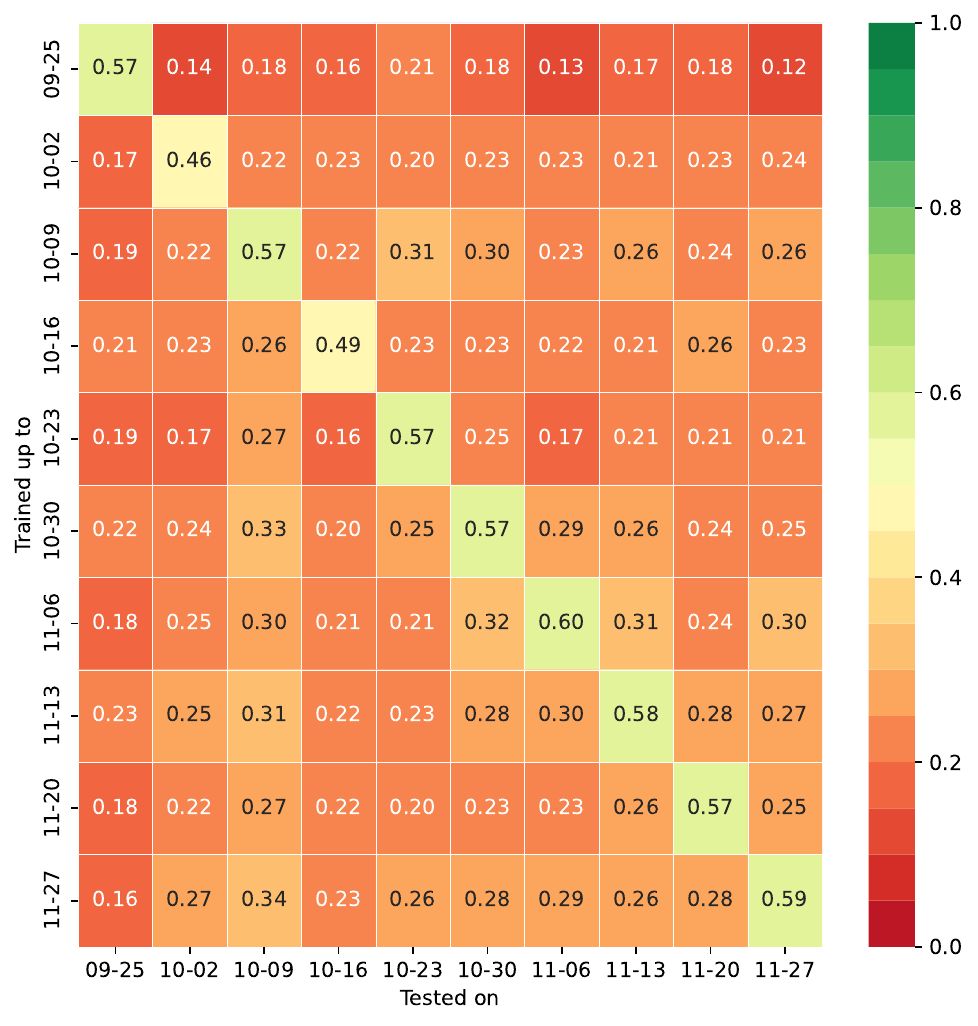}
         \caption{  MLP ($2$-hop)}
         \label{HeatmapMLP2_2022}
     \end{subfigure}  
        \caption{Accuracies for snapshots trained from September to December 2022. 
        Graph-MLP ($2$-hop), GCN ($2$-hop), and GCN ($2$-hop and edges) are omitted because they are similar to the values of the  MLP ($2$-hop).}
        \label{fig:heatmaps2022}
\end{figure}

The classification accuracies for the snapshots in 2012 and 2022 are shown in Figures~\ref{fig:heatmaps2012} and~\ref{fig:heatmaps2022}.
In general, the networks perform best on task $i$ after they were trained up to snapshot $i$.

% for 2012
Regarding 2012, the performance of the MLP for $\MACone$ is much higher than for $\MACtwo$.
The difference between these two MLPs is between $0.25$ and $0.41$. 
Furthermore, the best performance of a network is observed when tested on the first snapshot independent of the training snapshots.
For 2022, the results are similar to the experiments in 2012 but with overall lower accuracies.
The MLP accuracies for $\MACone$ are again much higher than those for $\MACtwo$. The difference is between $0.33$ and $0.53$. 
The best result for each snapshot~$i$ is achieved by the network that has been trained on snapshots $1$ to~$i$.
In contrast to 2012, the results on the first snapshot are not better than in the subsequent snapshots.

\begin{table*} 
\centering
\caption{Lifelong learning measures applied to the different networks and summaries of each year.}\label{tab:llmeasures}
\begin{tabular}{|l|l|l|rrrrrr|} 
 \hline 
 Year & Network & Type & ACC $\uparrow$ & BWT $\uparrow$ & FWT $\uparrow$ & $\Omega_\mathrm{base}$  $\uparrow$ & $\Omega_\mathrm{new}$  $\uparrow$ & $\Omega_\mathrm{all}$  $\uparrow$ \\ 
 \hline 
 2012 & MLP & $1$-hop & $0.739$ & $-0.101$ & $-0.062$ & $0.898$ & $0.829$ & $0.817$\\ 
   2012 & MLP & $2$-hop & $0.392$ & $-0.087$ & $-0.045$ & $0.827$ & $0.462$ & $0.673$\\ 
     2012 & Graph-MLP & $2$-hop & $0.420$ & $-0.099$ & $-0.049$ & $0.831$ & $0.499$ & $0.675$\\ 
      2012 & GCN & $2$-hop & $0.376$ & $-0.087$ & $-0.043$ & $0.824$ & $0.445$ & $0.672$\\ 
       2012 & GCN & $2$-hop and edges & $0.348$ & $-0.098$ & $-0.042$ & $0.850$ & $0.429$ & $0.690$\\ 
\hline
        2022 & MLP & $1$-hop & $0.701$ & $-0.263$ & $-0.275$ & $0.715$ & $0.938$ & $0.654$\\  
   2022 & MLP & $2$-hop & $0.296$ & $-0.289$ & $-0.312$ & $0.430$ & $0.555$ & $0.425$\\ 
     2022 & Graph-MLP & $2$-hop & $0.290$ & $-0.320$ & $-0.311$ & $0.437$ & $0.573$ & $0.426$\\ 
      2022 & GCN & $2$-hop &  $0.298$ & $-0.276$ & $-0.311$ & $0.426$ & $0.543$ & $0.421$\\  
      2022 & GCN & $2$-hop and edges &  $0.269$ & $-0.273$ & $-0.282$ & $0.431$ & $0.514$ & $0.426$\\ 
\hline
  \end{tabular}
\end{table*}

\paragraph{Lifelong Learning Measures}
\begin{figure}
\centering
\includegraphics[width=\linewidth]{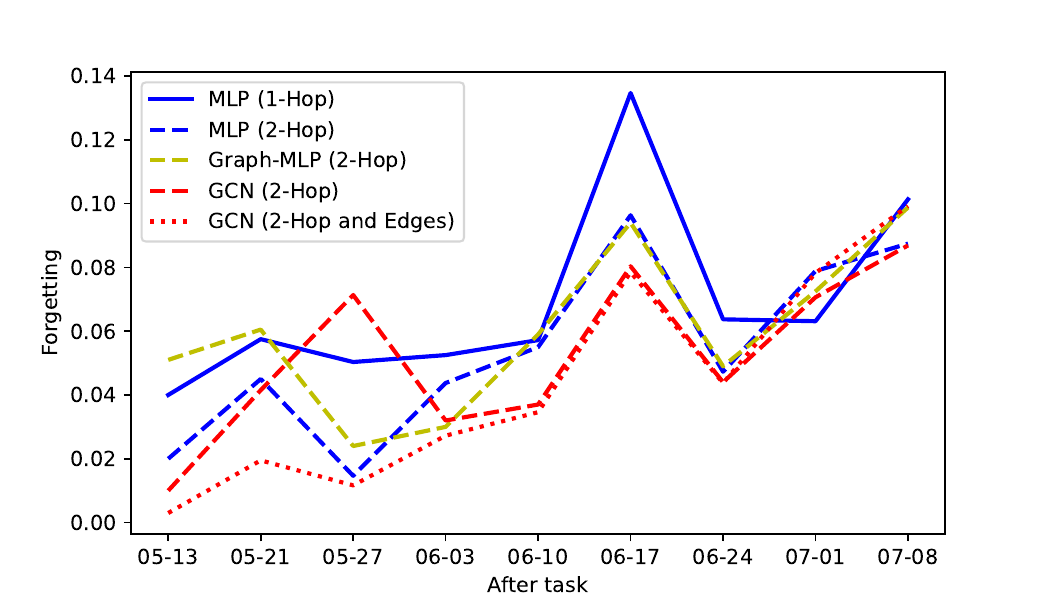}
            
         \caption{The progress of the forgetting in  2012.}
         \label{fig:forgetting_2012}
 \end{figure}  
\begin{figure}
\centering
\includegraphics[width=\linewidth]{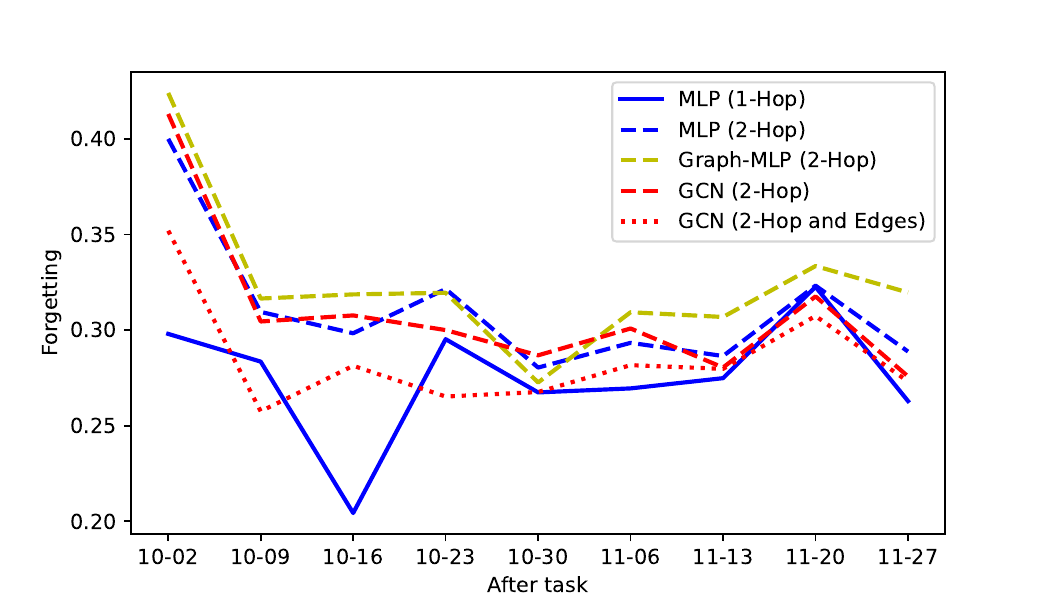}
         \caption{The progress of the forgetting in  2022.}
         \label{fig:forgetting_2022}
\end{figure}  
The lifelong learning measures are shown in Table~\ref{tab:llmeasures}. 
It shows that in 2012 and 2022, no network with any summary model or information achieved a positive forward or backward transfer in our setting.
For the other measures, the MLP applied on $\MACone$ achieves the best results, and the results for $\MACtwo$ are similar, independent of the network and the year.

The forgetting of the networks in 2012 is shown in Figure~\ref{fig:forgetting_2012}.
The MLP ($1$-hop) has the highest forgetting, with over $0.13$, after being trained on the 10~June snapshot. 
Overall, the networks forget more after being trained on this snapshot. 
After the networks are trained on the last snapshot in 2012, they have a similar forgetting value.

The forgetting of the networks in 2022, as shown in Figure~\ref{fig:forgetting_2022}, differs from the ones in 2012.
The highest forgetting for the $2$-hop networks is right in the beginning, and the lowest forgetting value is achieved by the  MLP ($1$-hop) after being trained on the 16~October snapshot.
In general, the forgetting values are higher in 2022 than in 2012.
Nevertheless, after the networks have been trained on the last snapshot, they have similar forgetting values.

\paragraph{Ten-year Time Warp}

\begin{table}
\centering
\caption{Accuracy after training on the task from September 25, 2022. Trained from scratch (no time warp, no TW) or initialized with weights from July 8, 2012 (TW).}
\label{tab:time warp}
\begin{tabular}{|l|l|c|c|} 
\hline
Neural network & Summary model & TW & no TW \\
  \hline
    MLP       & $1$-hop & $0.933$ & $0.930$ \\
    MLP       & $2$-hop & $0.581$ & $0.572$ \\ 
    Graph-MLP & $2$-hop & $0.605$ & $0.612$ \\ 
    GCN       & $2$-hop & $0.565$ & $0.572$ \\
    GCN       & $2$-hop and edges & $0.526$ & $0.515$ \\ 
    \hline
\end{tabular}

\end{table}

We show the results of reusing a neural network from 2012 to classify vertices in 2022 in Table~\ref{tab:time warp}. We call this setting a ten-year time warp (TW).
Retraining the last network from 2012 on the first snapshot in 2022 gives the same results as training a network from scratch in 2022. 
Table~\ref{tab:time warp} shows that the MLP for $\MACone$ performs a bit better in the time warp. 
For $\MACtwo$, the time warp makes the performance of the neural network better in two of four cases, and worse in the other two.
Reusing the last network from 2012 without retraining on the first 2022 snapshot gives very low accuracy -- always below $0.03$ (not shown in the table).

\section{Discussion}
\label{sec:discussion}

\paragraph{Vertex Classification}
Our experiments show that the problem of graph summarization by vertex classification is hard when applied to a temporal graph with changing and unseen classes.
The results from the snapshots in $2012$ and $2022$ (Figures~\ref{fig:heatmaps2012} and~\ref{fig:heatmaps2022}) show that the classification of EQCs for $\MACone$ is easier than for $\MACtwo$.
This is caused by the increased number of EQCs when more hops are considered in the summary model. 
Figures~\ref{AllEQCs2012m} and~\ref{AllEQCs2022m} show that there are more unique EQCs for $\MACtwo$ than for $\MACone$. 
The EQCs in $\MACtwo$ also change more rapidly.
The 2022 snapshots have more edges than the ones from 2012.
Overall, this reduces the performance. 
The more EQCs the network needs to learn and the more they change, the worse the network performs in general on all tasks.

%%% Forward transfer / backward
\paragraph{Forward and Backward Transfer}
We observe that no network has a positive forward or backward transfer between different snapshots. 
That means the networks perform better if they are trained on the task before being tested on it. 
The reason for the negative forward transfer is that new classes appear, and the reason for the negative backward transfer is that the neural network forgets classes.
The results (Figures~\ref{fig:heatmaps2012} and~ \ref{fig:heatmaps2022}) confirm this statement, since a network that has been trained on $\mathcal{T}_1, \dots, \mathcal{T}_i$ always achieves its best accuracy on~$\mathcal{T}_i$. 
This observation is confirmed by  $\Omega_\mathrm{new}$, the average of the accuracies on tasks after a network has been trained, 
which was higher than the other accuracies.
One explanation is that %the tasks the networks faced were very different, meaning that 
the EQCs of the snapshots change a lot, as can be seen in Figures~\ref{ChangesP2012-1m}-- \ref{ChangesP2022-2m}.

The $\Omega_\mathrm{base}$ values in 2012 show that for this set of snapshots, the networks perform well on the first snapshot at each time.
This is not the case for the networks for $\MACtwo$ in 2022, where the values are below~$0.5$. 
The networks do not perform as well as they did on the first snapshot. 
This shows that a network performs best on a task after being trained for it.

% Forgetting vs. Reuse
\paragraph{Forgetting and Reusing Networks}
The $ACC$ and the $\Omega_\mathrm{all}$ value of a network indicate how well a neural network performs on previous tasks. 
Figures~\ref{fig:forgetting_2012} and \ref{fig:forgetting_2022} show how much a network forgets over time.  
A higher $ACC$ and $\Omega_\mathrm{all}$  value do not necessarily mean a network forgets less. 
A good example is the MLP ($1$-hop) of 2012, which has a better $ACC$  and $\Omega_\mathrm{all}$  value than the $2$-hop models, but the forgetting rate is higher.

\paragraph{Using $1$-hop versus $2$-hop Information in a $2$-hop Summary Model}
We investigate whether the information provided from a $1$-hop neighborhood versus using the $2$-hop neighborhood is helpful. 
The measures show that the networks for $\MACtwo$, independent of the information used, performed similarly, and this held in both 2012 and 2022.
Graph-MLP and GCN perform similarly to MLP for $\MACtwo$, even though they have access to $2$-hop information for the classification task and MLP does not.
In summary, the results show that $2$-hop information does not improve the performance for $\MACtwo$.
While this seems to be counter-intuitive, an explanation is that for the $2$-hop summary, there are five to ten times more classes than in the $1$-hop summary since the number of paths grows exponentially with the number of hops (see Figures \ref{AllEQCs2012m} and~\ref{AllEQCs2022m}). 
This results in many small EQCs, \ie many classes with low support that are hard to train on.

\paragraph{Warm versus Cold Restarts}
\citet{galke-neuralnetworks-journal} differentiate between warm restart and cold restart in the context of lifelong learning. 
A warm restart is reusing network parameters from previous tasks, and retraining from scratch is called a cold restart. 
Our time warp experiment (Table~\ref{tab:time warp}) does not show a clear benefit of a warm restart compared to a cold restart over a ten-year range. 
However, it shows that a warm restart does not hurt a network's performance.

\paragraph{Limitations and Threat to Validity}
The networks may perform better if we trained them for more iterations. 
However, when we increase the number of training iterations by an order of magnitude, we observe an improvement of only $2$~points on the classification accuracy.
Training more on the current task may also increase the forgetting of previous tasks. 
Furthermore, we use a hash function to compute the gold standard of the vertex labels, \ie the EQCs.
There could be collisions in the hash function.
However, we chose the hash such that the probability is very low following \citet{DBLP:conf/dsaa/BlasiFHRS22}.

\section{Conclusion}
\label{sec:conclusion}
We show that graph summarization for temporal graphs using a neural network is more challenging than the equivalent task on static graphs.
The performance of the vertex classification in a temporal graph depends on the number of classes and the changes between timesteps.
A problem with the summarization is that many new EQCs appear, and known ones disappear over time. 
We observe forgetting in lifelong graph summarization by vertex classification. 
We also observe that the performance of a network should not only be measured by the accuracy of the current task but also by other measures, like forward and backward transfer, which should be taken into account.
The measures show that the networks for $\MACtwo$ perform similarly. 
With the time warp experiments, we see that reusing the network parameters does not necessarily improve lifelong learning measures.

\section*{Acknowledgements}
This paper is the result of the first author's master's project in data science.
This research is co-funded by the CodeInspector project (No. 504226141) of the DFG, German Research Foundation, and the 2LIKE project by the German Federal Ministry of Education and Research (BMBF) and the Ministry of Science, Research and the Arts Baden-Württemberg within the funding line Artificial Intelligence in Higher Education.
The authors acknowledge support by the state of Baden-Württemberg through bwHPC.
\bibliographystyle{IEEEtranN}
{\footnotesize\bibliography{main}}

\clearpage
\onecolumn 

\appendix\label{sec:appendix}

\subsection{Hyperparameter Optimization}
\label{sec:hyp}

The results of our hyperparameter optimization are shown in Table~\ref{tab:val}.
Since the search is very expensive, we first perform a grid search with each network and configuration for $46$ iterations. 
Subsequently, we narrowed the search and chose the three best configurations of a network for each summary.
We validate every neural network for $751$ more iterations, except MLP when using the graph summary model $M_{AC1}$. 
For MLP with $M_{AC1}$, we use only $121$ iterations because its batches contain more samples than for the graph neural networks. 
The results show that training for more iterations does not improve the performance.

\begin{table}[H]
    \centering

    \begin{tabular}{|l|l|l|l|l|l|l|rrrrr|}\hline
Network& Model &LS	&	LR	&	D	&	$\alpha	$&	$\tau$	&	Acc(1)	&	Acc(2)	&	Acc(3)	&	Acc(4)		&	Acc(5)	\\ \hline 
MLP &$M_{AC1}$ &	1,024	&	0.1	&	0.5	&	& &	0.8481	&	0.8380	&	0.8357	&	0.8297	&	0.8440	\\ 	
\textbf{MLP} &\textbf{$M_{AC1}$} &	\textbf{1,024}&	\textbf{0.01}	&	\textbf{0.5}&	& &	\textbf{0.7969	}&	\textbf{0.8398	}&	\textbf{0.8499	}&	\textbf{0.8496}	&	\textbf{0.8532	}	\\ 	
MLP &$M_{AC1}$ &	 1,024	&	0.1	&	0.2	&	& &	 0.8378	& 	0.8367	&	0.8293	&	0.8366	&		0.8367	\\ 	
\hline 
\textbf{MLP} &\textbf{$M_{AC2}$} &	\textbf{1,024}&	\textbf{0.01}&	\textbf{0.5	}&	& &		\textbf{0.5758}&		\textbf{0.5775}	&	\textbf{0.5807}	&	\textbf{0.5832	}&	\textbf{0.5752}	\\ 	
MLP &$M_{AC2}$ &	1,024	&	0.1	&	0.5	&		& &	 0.5479	&	0.5562	&	0.5524	&	0.5548	&	0.5458\\ 	
MLP &$M_{AC2}$ &	1,024	&	0.1	&	0.2	&	& &		0.5539	&	0.5633	&	0.5546	&		0.5543	&	0.5518		\\   	\hline 
 GCN &$M_{AC2}$ &	64	&	0.1	&	0.2	&	& &	0.5654	&	0.5589	&	0.5714	&	0.5575	&		0.5621	\\ 	
 GCN &$M_{AC2}$ &	32	&	0.1	&	0.2	&	& &	0.5686	&	0.5639	&	0.5518	&	0.5559	&	0.5588	\\ 	
\textbf{GCN} &\textbf{$M_{AC2}$} &	\textbf{64	}&	\textbf{0.1	}&	\textbf{0	}&	& &		\textbf{0.5736	}&	\textbf{0.5797	}&	\textbf{0.5720	}&	\textbf{0.5788	}&	\textbf{0.5686}	\\ 	\hline 
\textbf{GCN (Edges)} &\textbf{$M_{AC2}$} &		\textbf{32	}&	\textbf{0.1	}&	\textbf{0}	&	& &		\textbf{0.5558}	&		\textbf{0.5665	}&	\textbf{0.5550	}&	\textbf{0.5656	}	&	\textbf{0.5540}	\\ 	
GCN (Edges)&$M_{AC2}$ &	64	&	0.1	&	0	&		& &		0.5409	&	0.5566	&	0.5481	&	0.5389	&	0.5352		\\ 	
GCN (Edges)&$M_{AC2}$ &	64	&	0.1	&	0.5	&	& &		0.5350	&	0.5437	&	0.5223	&	0.5350	&		0.5342	\\ 	\hline 
\textbf{Graph-MLP} &\textbf{$M_{AC2}$} &	\textbf{64	}&	\textbf{0.01	}&	\textbf{0.2	}&	\textbf{1}	&	\textbf{2	}&	\textbf{0.6184	}&	\textbf{0.6250	}&		\textbf{0.6280	}&	\textbf{0.6180	}	&	\textbf{0.6286	}	\\ 	
Graph-MLP &$M_{AC2}$ &	64	&	0.01	&	0	&	1	&	0.5	&	0.6222	&	0.6294		&	0.6253		&	0.6257	&	0.6269	\\ 	
Graph-MLP &$M_{AC2}$ &	64	&	0.01	&	0	&	10	&	0.5	&	0.6195	&	0.6228		&		0.6210	&	0.6246	&	0.6258	\\ 	\hline 
 \end{tabular}
  \caption{
  Extensive search for parameters for the networks. Abbreviations: Layer Size (LS), Learning Rate (LR), and  Dropout (D)). For MLP ($M_{AC1}$), the accuracy Acc($n$) is measured after the network has been trained for $1+n\cdot 24$ iterations and for the other networks after they have been trained for  $1+n\cdot 150$ iterations. Best networks are in bold.}\label{tab:val}
\end{table}
\subsection{Accuracies}
\begin{figure*}
     \centering
     \hfill
     \begin{subfigure}[b]{0.3\textwidth}
         \centering
         \includegraphics[width=\textwidth]{images/accuracy_mlp_1-1-cropped.pdf}
         \caption{ MLP ($1$-hop)}
         \label{HeatmapMLP1_2012}
     \end{subfigure} 
     \begin{subfigure}[b]{0.3\textwidth}
         \centering
         \includegraphics[width=\textwidth]{images/accuracy_mlp_2-1-cropped.pdf}
         \caption{ MLP ($2$-hop)}
         \label{HeatmapMLP2_2012}
     \end{subfigure} 
     \begin{subfigure}[b]{0.3\textwidth}
         \centering
         \includegraphics[width=\textwidth]{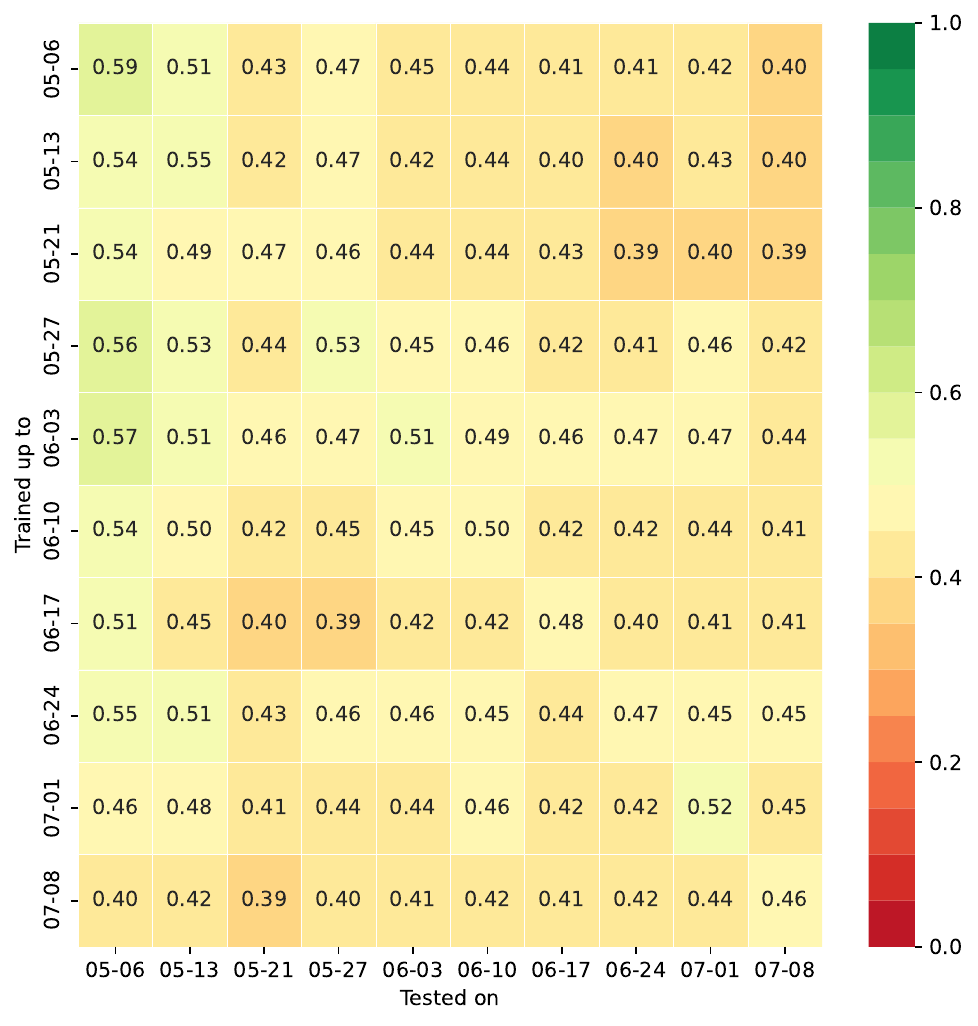}
         \caption{  Graph-MLP ($2$-hop)}
         \label{HeatmapGraphMLP2_2012}
     \end{subfigure} 
     \begin{subfigure}[b]{0.3\textwidth}
         \centering
         \includegraphics[width=\textwidth]{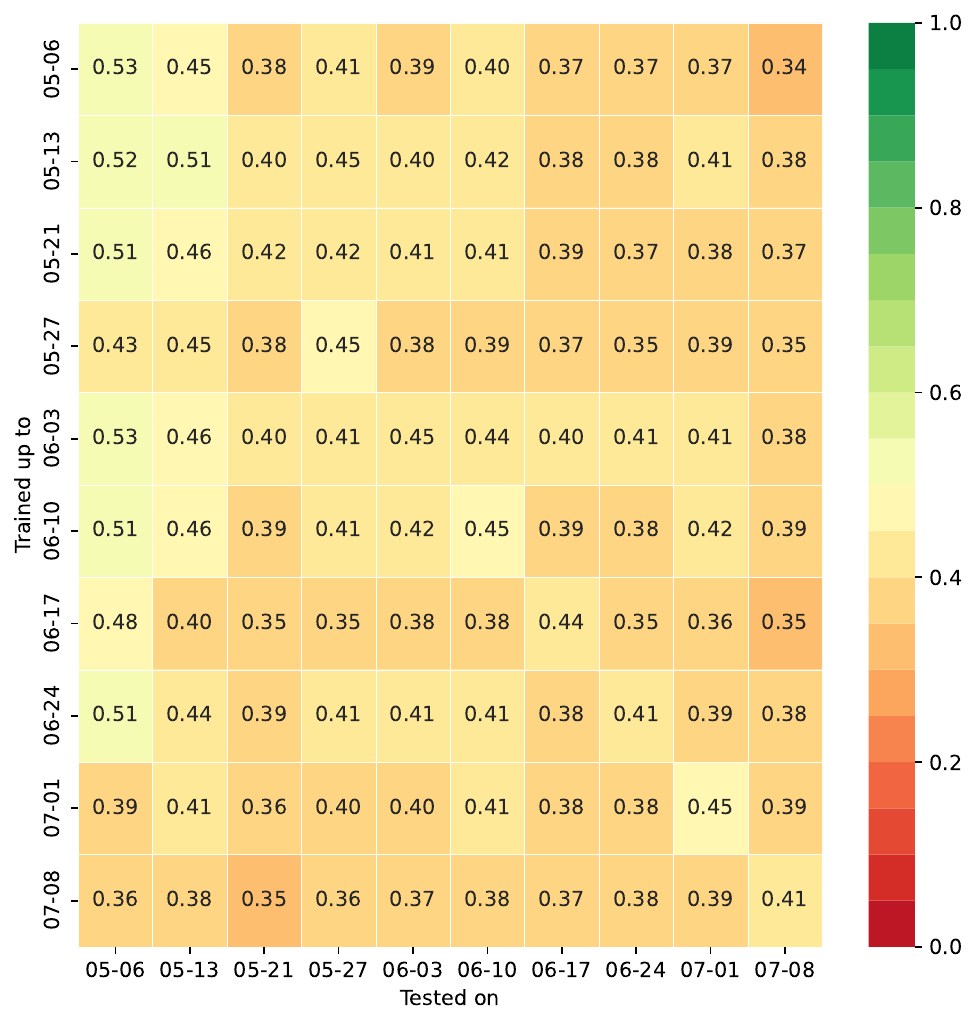}
          \caption{ GCN ($2$-hop)}
         \label{HeatmapGraphSAINT2_2012}
     \end{subfigure} 
     \begin{subfigure}[b]{0.3\textwidth}
         \centering
         \includegraphics[width=\textwidth]{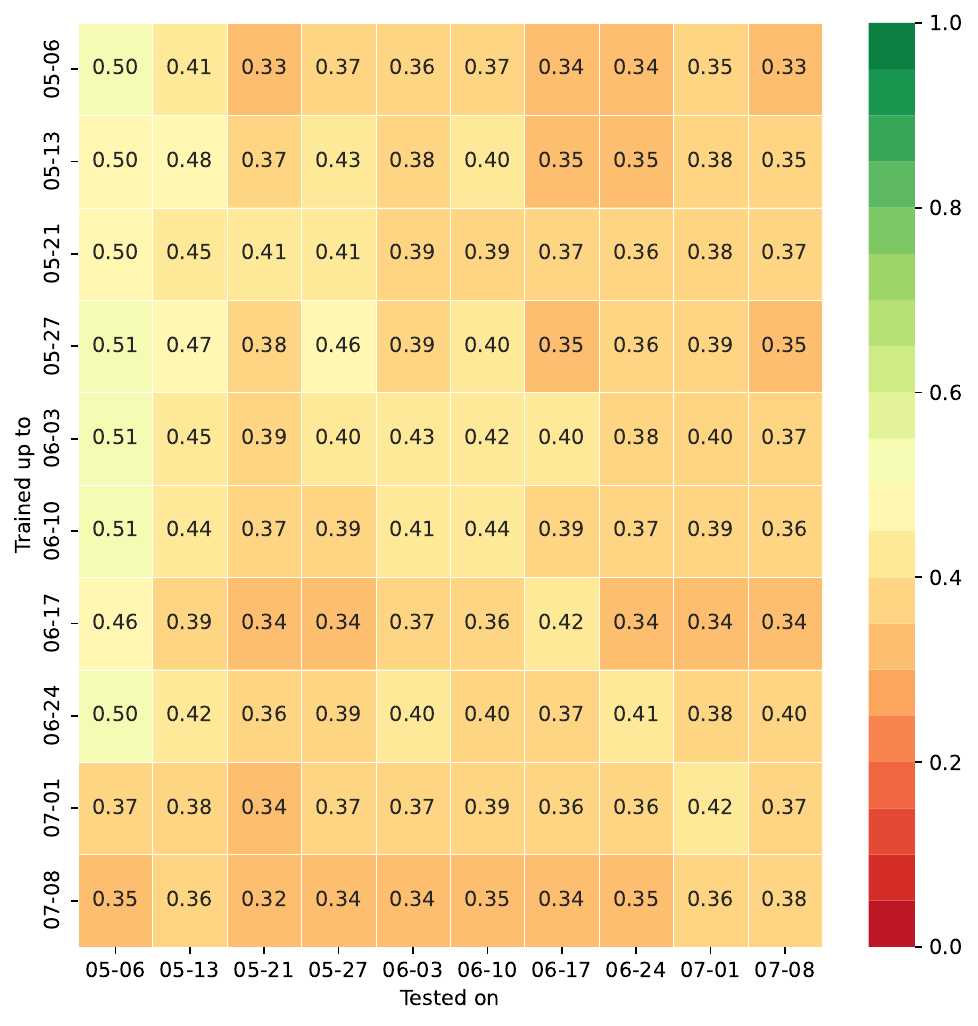}
         \caption{ GCN ($2$-hop and edges)}
         \label{HeatmapGraphSAINT3_2012}
     \end{subfigure}  \hfill
        \caption{Accuracies for snapshots trained from May to July 2012.}
        \label{fig:heatmaps2012A}
\end{figure*}

\begin{figure*}
     \centering
      \hfill
     \begin{subfigure}[b]{0.3\textwidth}
         \centering
         \includegraphics[width=\textwidth]{images/accuracy_mlp_1-2-cropped.pdf}
         \caption{  MLP ($1$-hop)}
         \label{HeatmapMLP1_2022}
     \end{subfigure}  
     \begin{subfigure}[b]{0.3\textwidth}
         \centering
         \includegraphics[width=\textwidth]{images/accuracy_mlp_2-2-cropped.pdf}
         \caption{  MLP ($2$-hop)}
         \label{HeatmapMLP2_2022}
     \end{subfigure}  
     \begin{subfigure}[b]{0.3\textwidth}
         \centering
         \includegraphics[width=\textwidth]{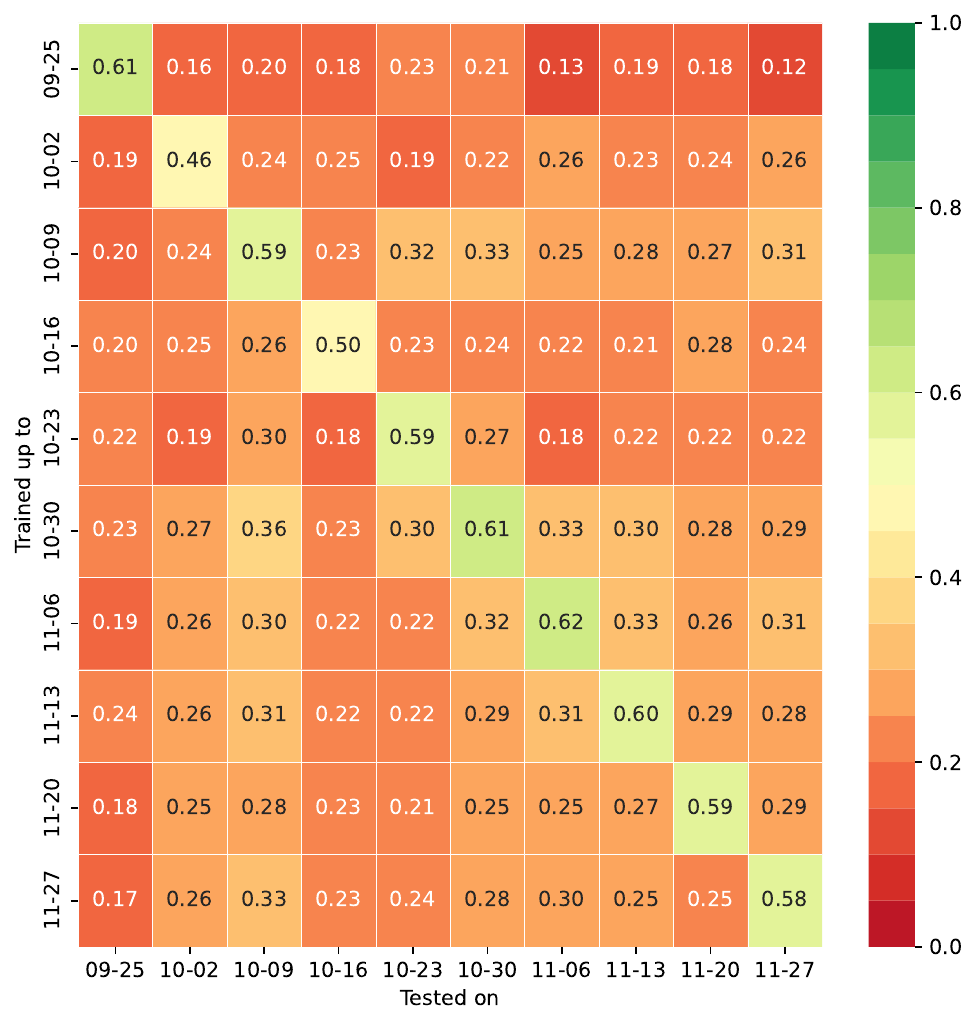}
         \caption{  Graph-MLP ($2$-hop)}
         \label{HeatmapGraphMLP2_2022}
     \end{subfigure} 
     \begin{subfigure}[b]{0.3\textwidth}
         \centering
         \includegraphics[width=\textwidth]{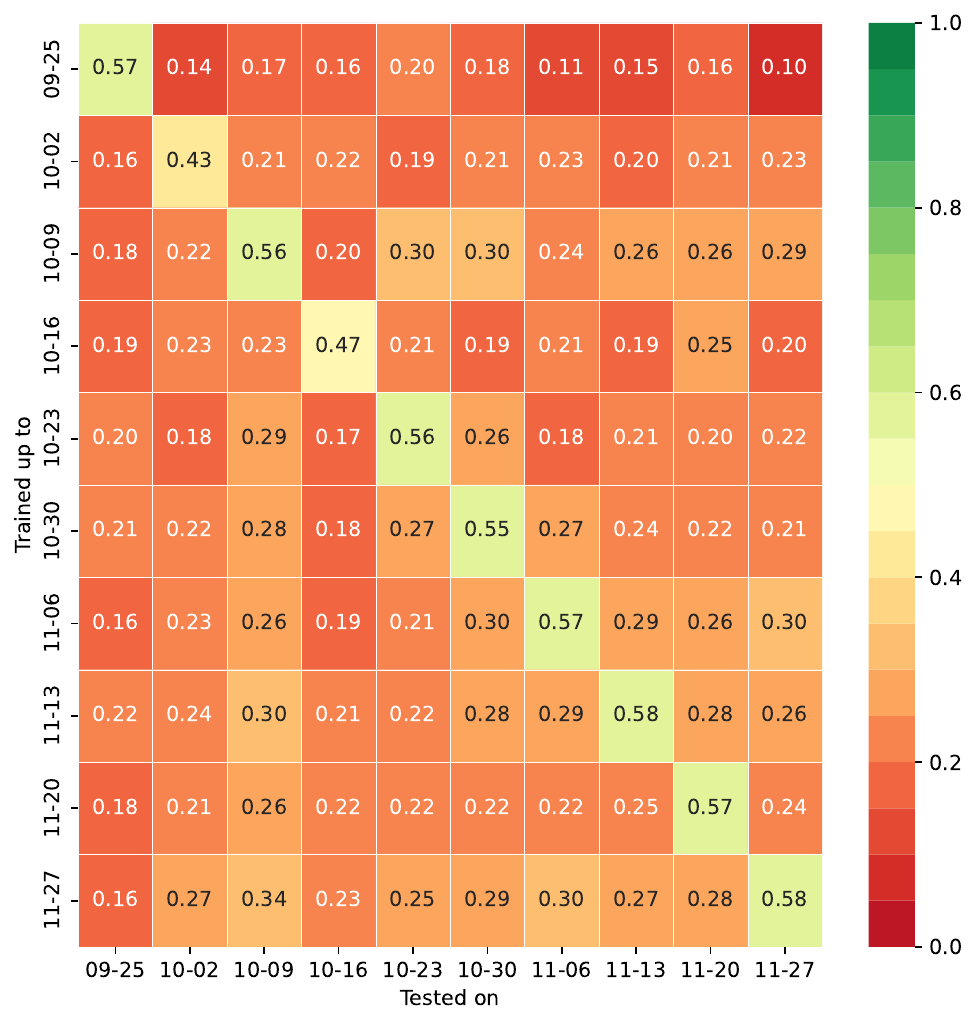}
          \caption{GCN ($2$-hop)}
         \label{HeatmapGraphSAINT2_2022}
     \end{subfigure}  
     \begin{subfigure}[b]{0.3\textwidth}
         \centering
         \includegraphics[width=\textwidth]{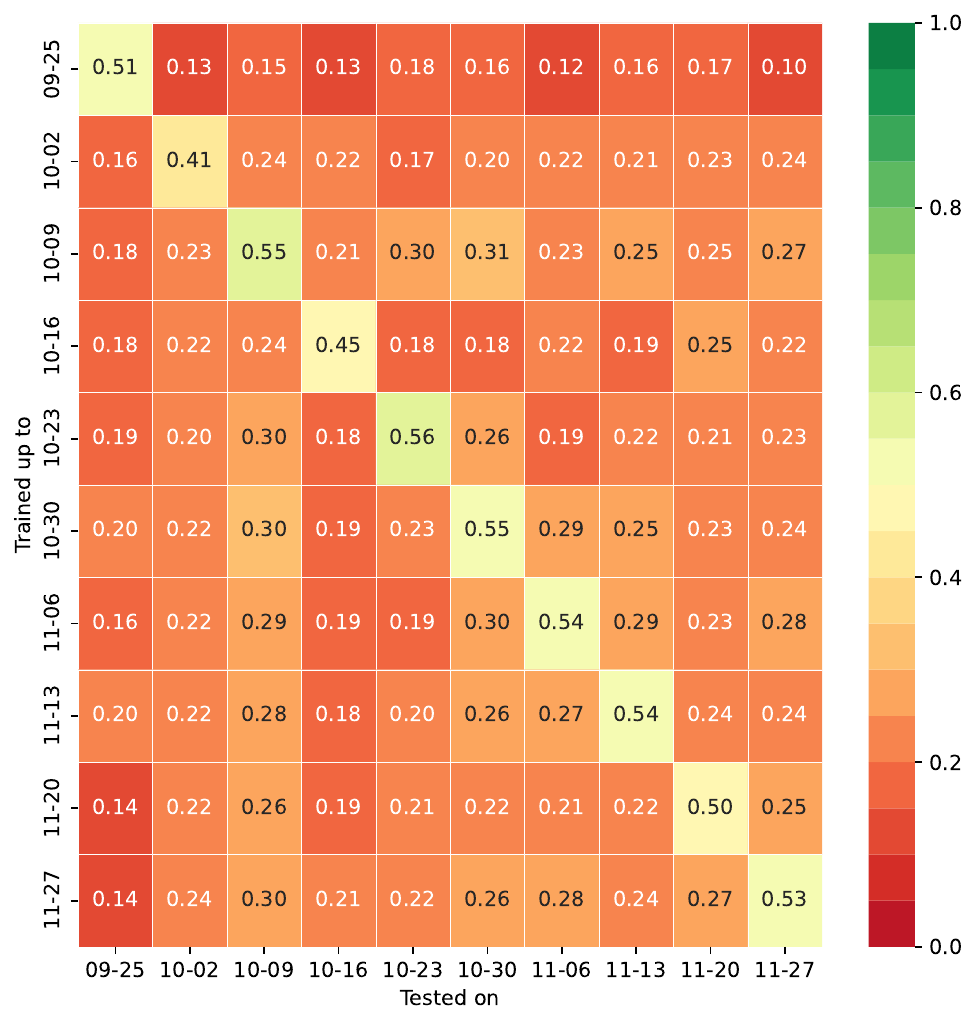}
         \caption{GCN ($2$-hop and edges)}
         \label{HeatmapGraphSAINT3_2022}
     \end{subfigure}  \hfill
        \caption{Accuracies for snapshots trained from September to December 2022.}
        \label{fig:heatmaps2022A}
\end{figure*}

The accuracies for all models for 2012 and 2022 are shown in Figure~\ref{fig:heatmaps2012A} and~\ref{fig:heatmaps2022A}. 

\subsection{Graph Summary Measures}
\label{sec:Measures}
To analyze the graphs and summaries more extensively, we exploit additional measures.
We divide these measures into three categories.
The first category is the unary measures. The second is binary measures, and the last is meta-measures.

\subsubsection{Unary Measures}
Unary measures only consider a summary $S_t$ at time $t$. 
The features of each vertex in the summary are the hash value that indicates its EQC, the degree, and the members of the EQC in the original graph. For each summary $S_t$, we have a function $ext_t()$ that returns for each vertex in the summary the number of vertices extensions in the original graph  $G_t$.
For the vertices, i.e., EQCs in the summary,  $S_t$, we calculate the average size of EQCs as
\begin{equation}\delta^{(1)}_{AvgSize}(S_t) = \frac{\sum_{q \in V^S_t}ext_t(q)}{|V^S_t|},\end{equation}
and the average number of edges/attributes per EQC as 
\begin{equation}\delta^{(1)}_{AvgE}(S_t)=\frac{|E^S_t|}{|V^S_t|}.\end{equation}
   We also calculate distributions over the number of attributes of an EQC, over the number of members of an EQC, and over the number of usages of a predicate.

\begin{figure}[H]
     \centering
     \begin{subfigure}[b]{0.24\textwidth}
         \centering
         \includegraphics[width=\textwidth]{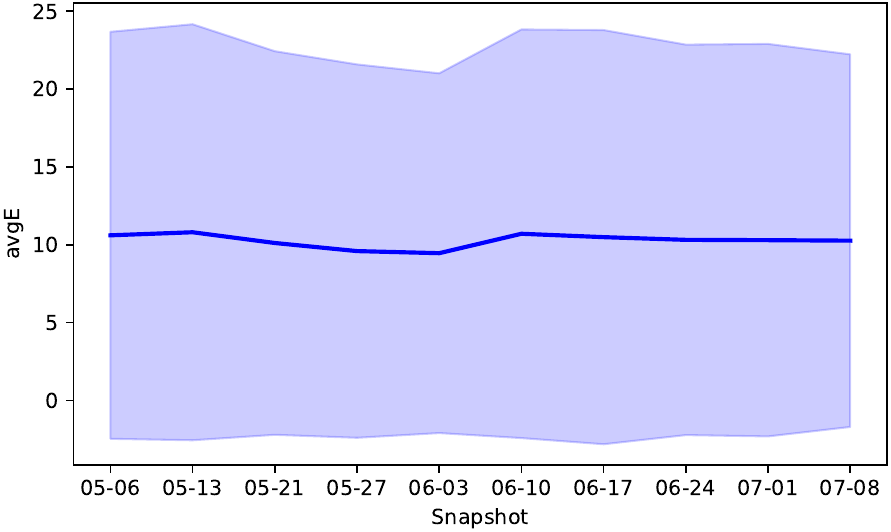}
         \caption{ Average number of attributes per EQC in 2012 ($1$-hop).}
         \label{AvgA2012-1}
     \end{subfigure}  \hfill
     \begin{subfigure}[b]{0.24\textwidth}
         \centering
         \includegraphics[width=\textwidth]{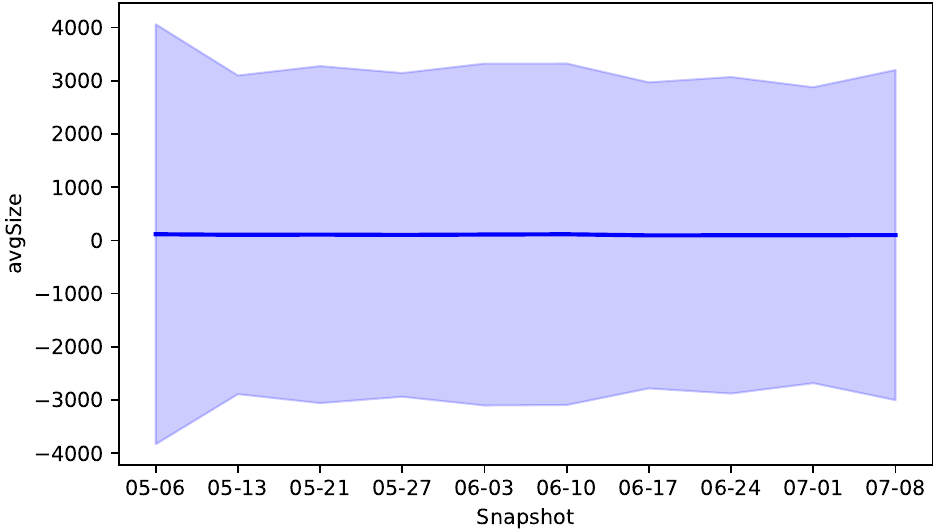}
         \caption{ Average number of members per EQC in 2012 ($1$-hop).}
         \label{AvgM2012-1}
     \end{subfigure}  \hfill
     \begin{subfigure}[b]{0.24\textwidth}
         \centering
         \includegraphics[width=\textwidth]{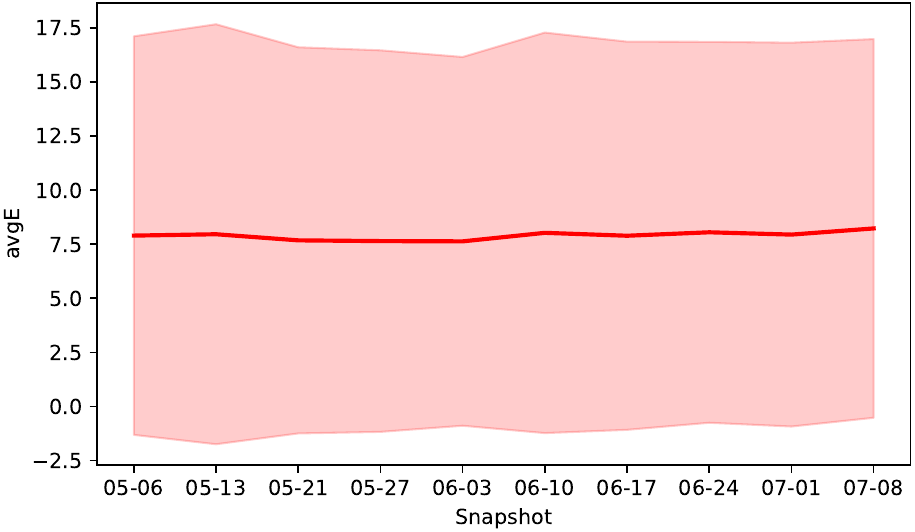}
         \caption{Average number of attributes per EQC in 2012 ($2$-hop).}
         \label{AvgA2012-2}
     \end{subfigure}  \hfill
     \begin{subfigure}[b]{0.24\textwidth}
         \centering
         \includegraphics[width=\textwidth]{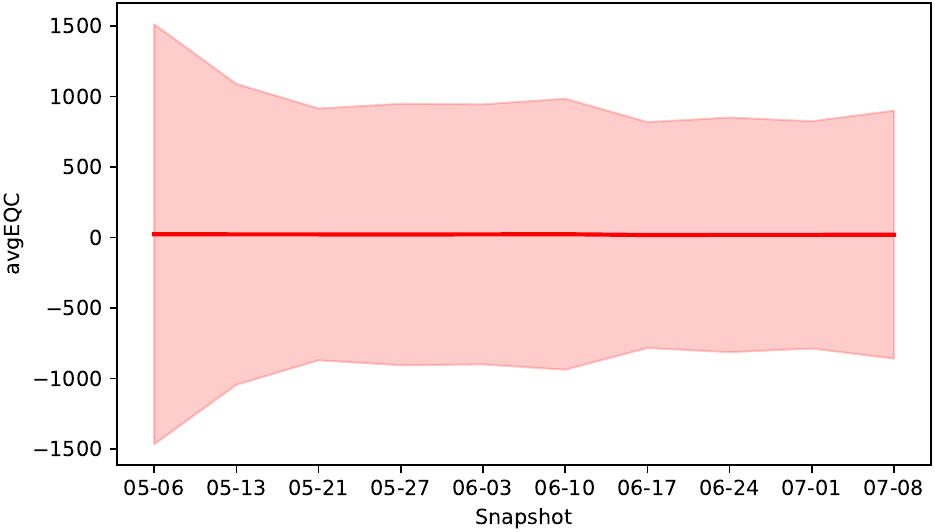}
         \caption{ Average number of members per EQC in 2012 ($1$-hop).}
         \label{AvgM2012-2}
     \end{subfigure}  \hfill
     \begin{subfigure}[b]{0.24\textwidth}
         \centering
         \includegraphics[width=\textwidth]{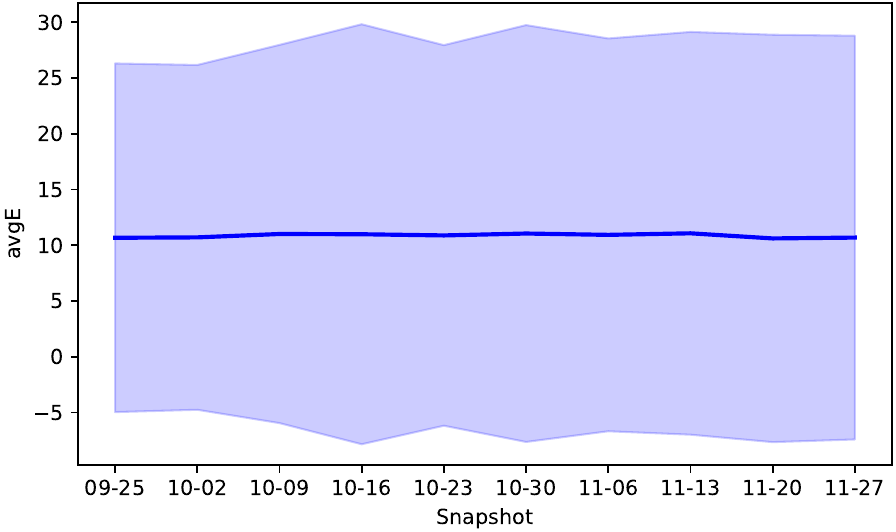}
           \caption{ Average number of attributes per EQC in 2022 ($1$-hop).}
         \label{AvgA2022-1}
     \end{subfigure}  \hfill
     \begin{subfigure}[b]{0.24\textwidth}
         \centering
         \includegraphics[width=\textwidth]{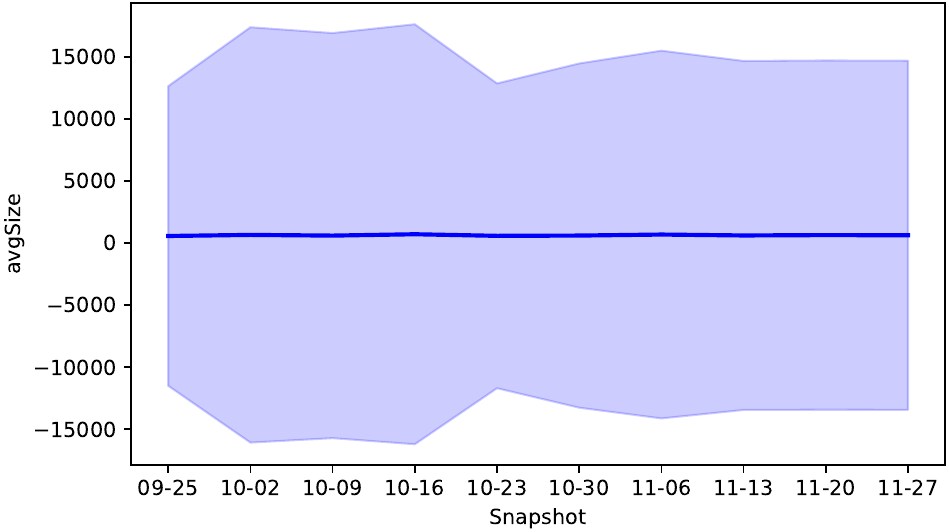}
         \caption{Average number of members per EQC in 2022 ($1$-hop).}
         \label{AvgM2022-1}
     \end{subfigure}  \hfill
     \begin{subfigure}[b]{0.24\textwidth}
         \centering
         \includegraphics[width=\textwidth]{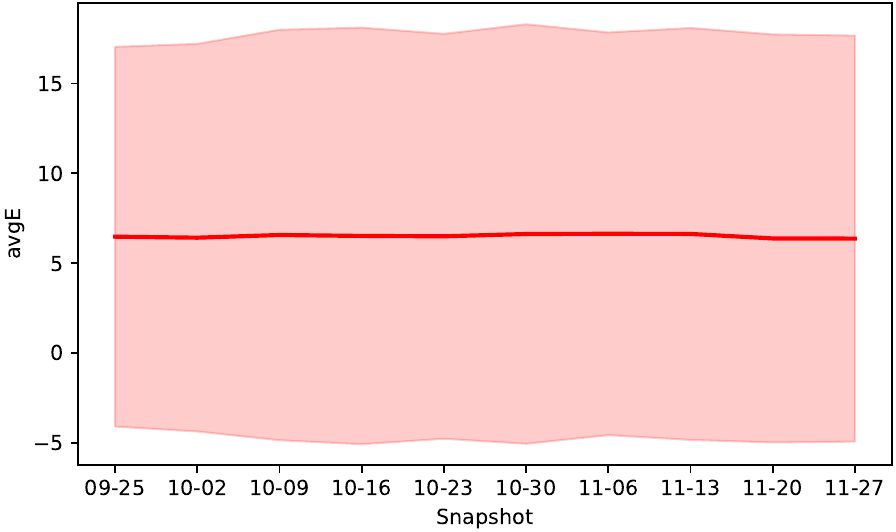}
         \caption{ Average number of attributes per EQC in 2022 ($2$-hop).}
         \label{AvgA2022-2}
     \end{subfigure}  \hfill
     \begin{subfigure}[b]{0.24\textwidth}
         \centering
         \includegraphics[width=\textwidth]{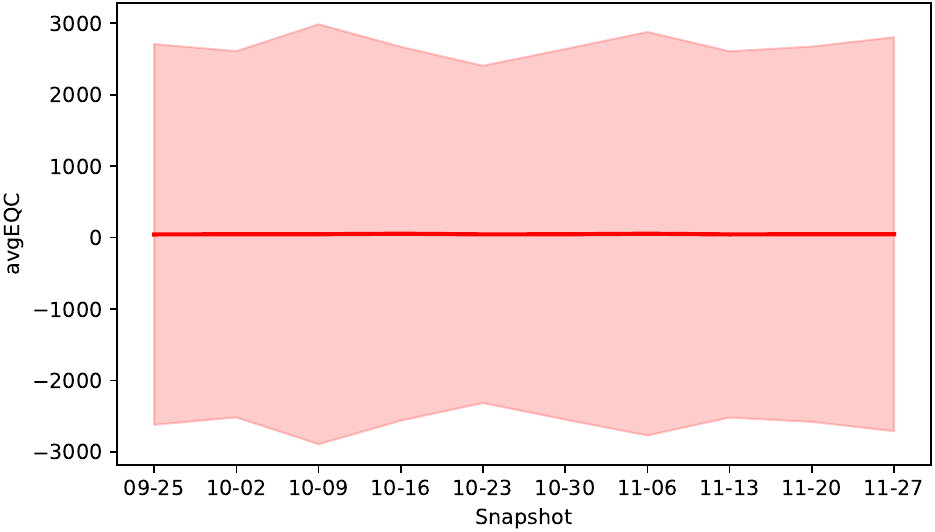}
         \caption{Average number of members per EQC in 2022 ($2$-hop).}
         \label{AvgM2022-2}
     \end{subfigure}  \hfill
         \caption{Average numbers for attributes (AvgE) and members (AvgSize) per EQC in the snapshots 2012 and 2022.}
        \label{fig:deviation}
\end{figure}

\begin{figure}[H]
     \centering
     \begin{subfigure}[b]{0.3\textwidth}
         \centering
         \includegraphics[width=\textwidth]{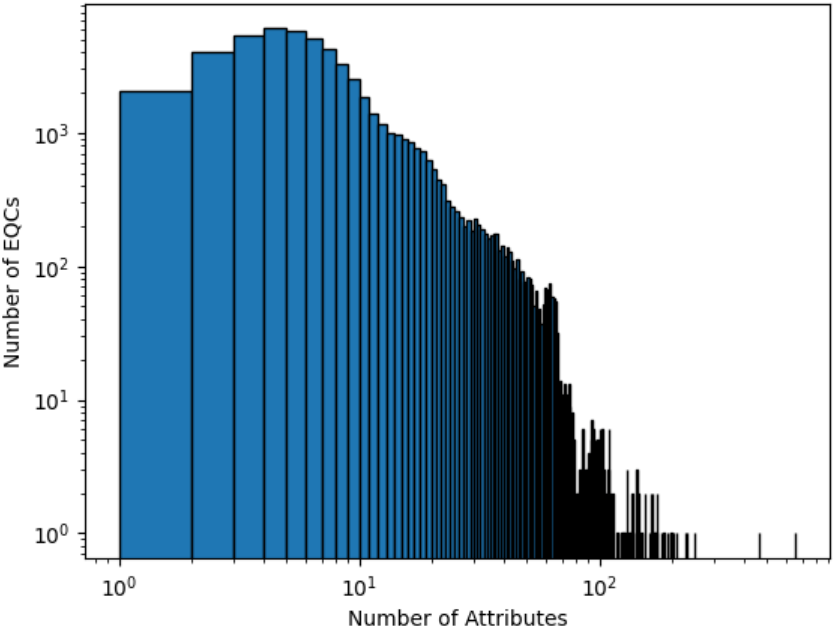}
         \caption{Frequency of number of attributes per EQC (in descending order) for the $1$-hop attribute collection.}
     \end{subfigure}  \hfill
     \begin{subfigure}[b]{0.3\textwidth}
         \centering
         \includegraphics[width=\textwidth]{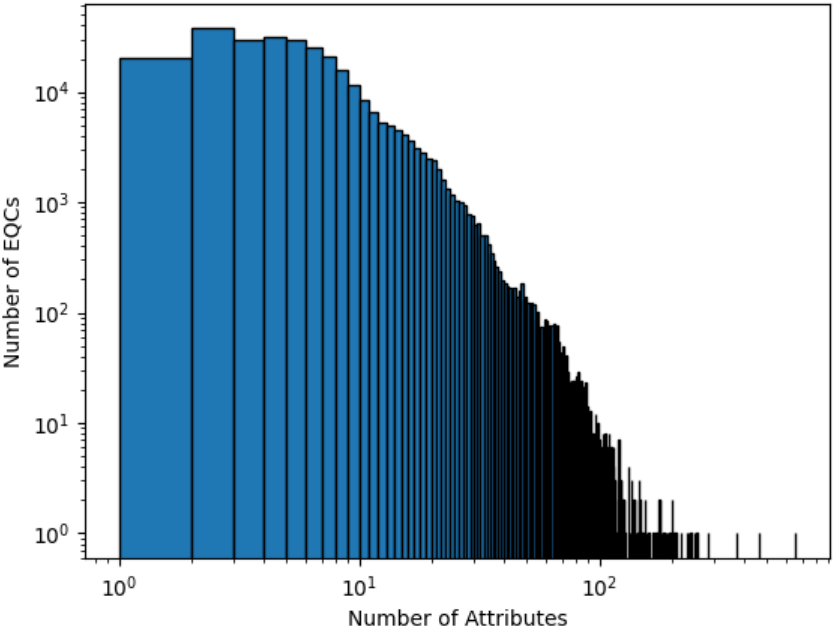}
         \caption{Frequency of number of attributes per EQC (in descending order) for the $2$-hop attribute collection.}
     \end{subfigure}  \hfill
     \begin{subfigure}[b]{0.3\textwidth}
         \centering
         \includegraphics[width=\textwidth]{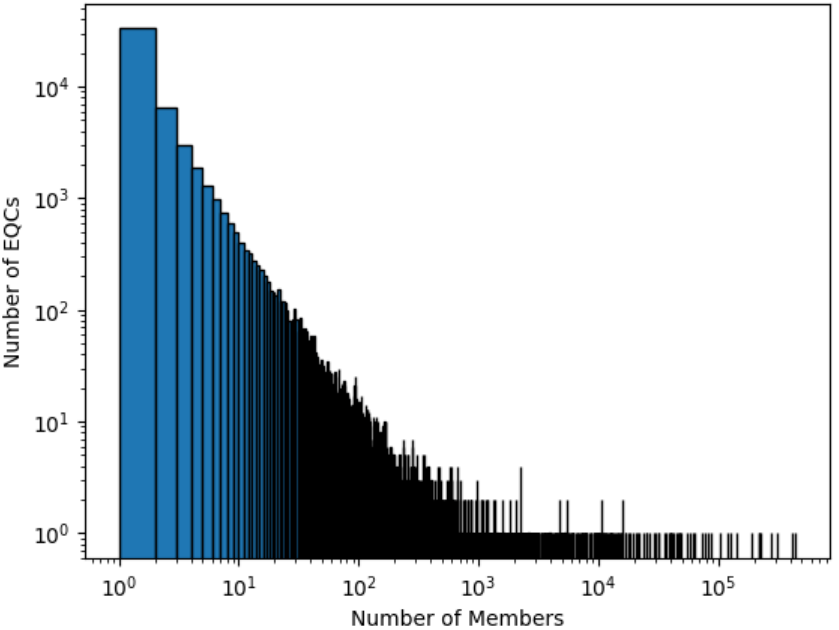}
         \caption{Frequency of number of members of an EQCs (in descending order) for the $1$-hop attribute collection.}
     \end{subfigure}  \hfill
     \begin{subfigure}[b]{0.3\textwidth}
         \centering
         \includegraphics[width=\textwidth]{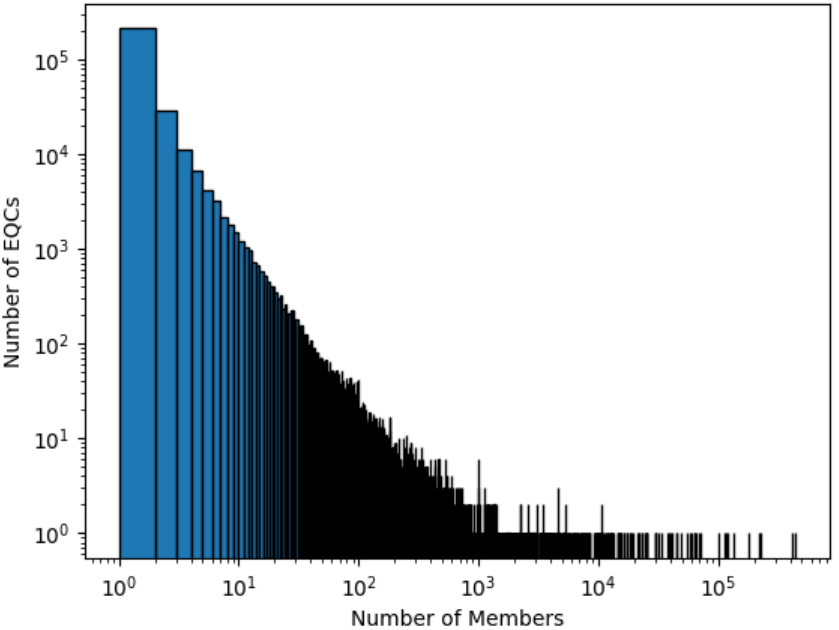}
         \caption{Frequency of number of members of an EQCs (in descending order) for the $2$-hop attribute collection.}
     \end{subfigure} 
     \begin{subfigure}[b]{0.3\textwidth}
         \centering
         \includegraphics[width=\textwidth]{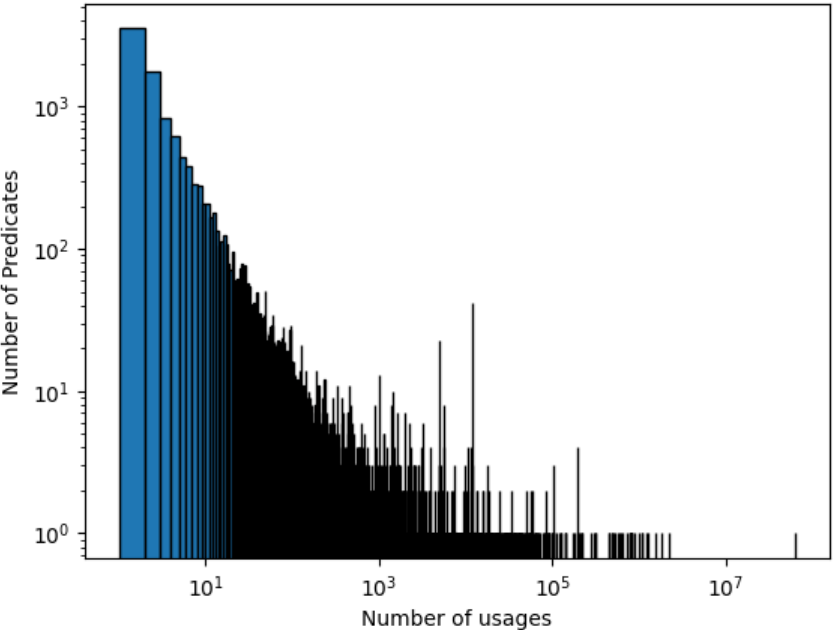}
         \caption{Frequency how often a predicate is used in the original graph (in descending order).}
     \end{subfigure} \hfill
        \caption{Distributions measures on the summary of 2012-05-06}
        \label{fig:dist2012-05}
\end{figure}

\begin{figure}[H]
     \centering
     \begin{subfigure}[b]{0.3\textwidth}
         \centering
         \includegraphics[width=\textwidth]{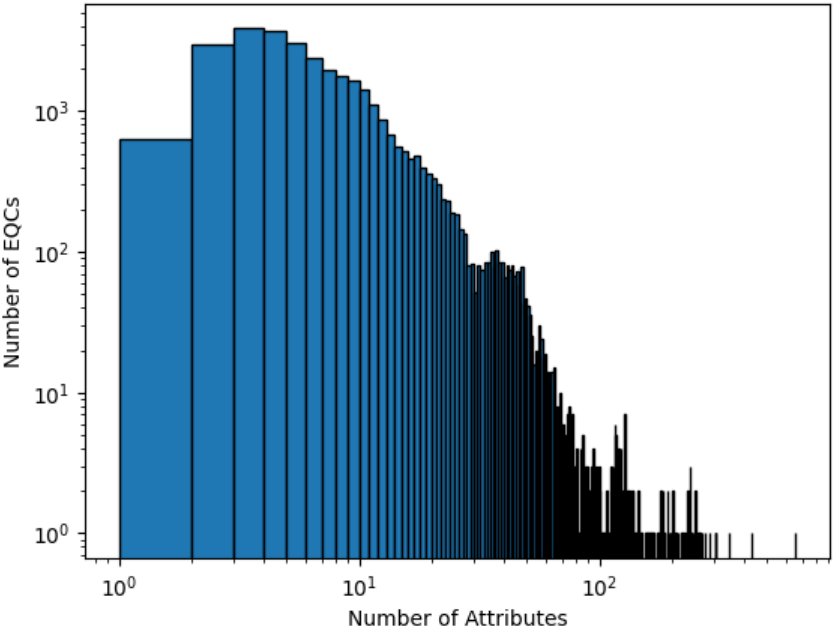}
         \caption{Frequency of number of attributes per EQC (in descending order) for the $1$-hop attribute collection.}
     \end{subfigure}  \hfill
     \begin{subfigure}[b]{0.3\textwidth}
         \centering
         \includegraphics[width=\textwidth]{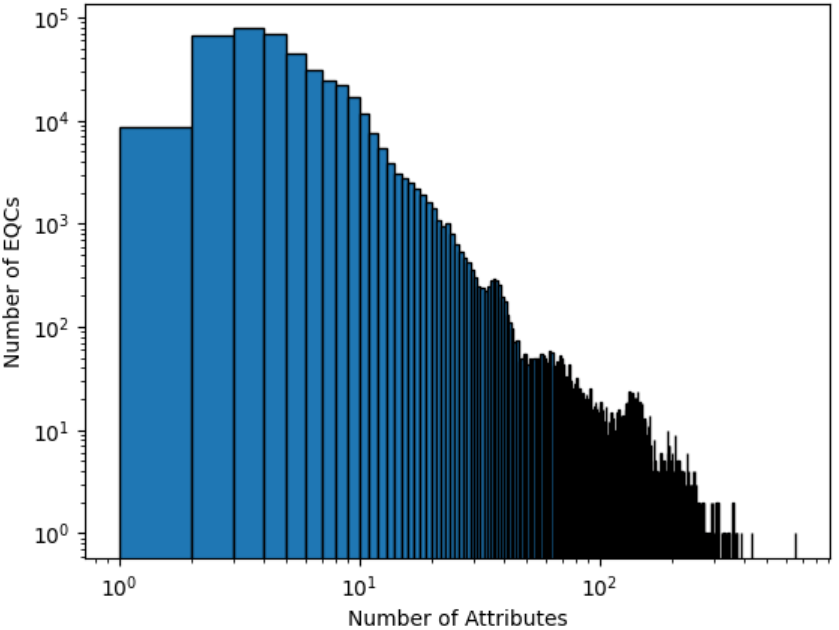}
         \caption{Frequency of number of attributes per EQC (in descending order) for the $2$-hop attribute collection.}
     \end{subfigure}  \hfill
     \begin{subfigure}[b]{0.3\textwidth}
         \centering
         \includegraphics[width=\textwidth]{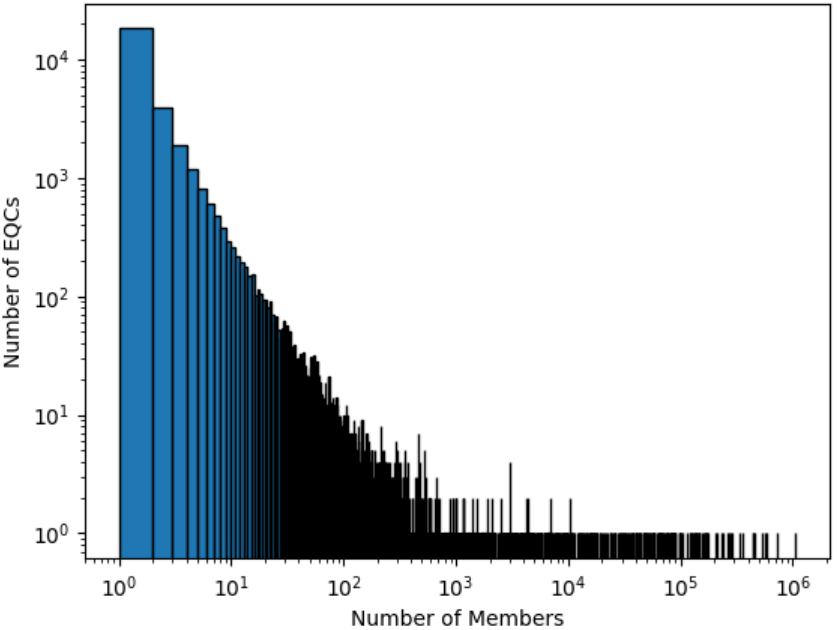}
         \caption{Frequency of number of members of an EQCs (in descending order) for the $1$-hop attribute collection.}
     \end{subfigure}  \hfill
     \begin{subfigure}[b]{0.3\textwidth}
         \centering
         \includegraphics[width=\textwidth]{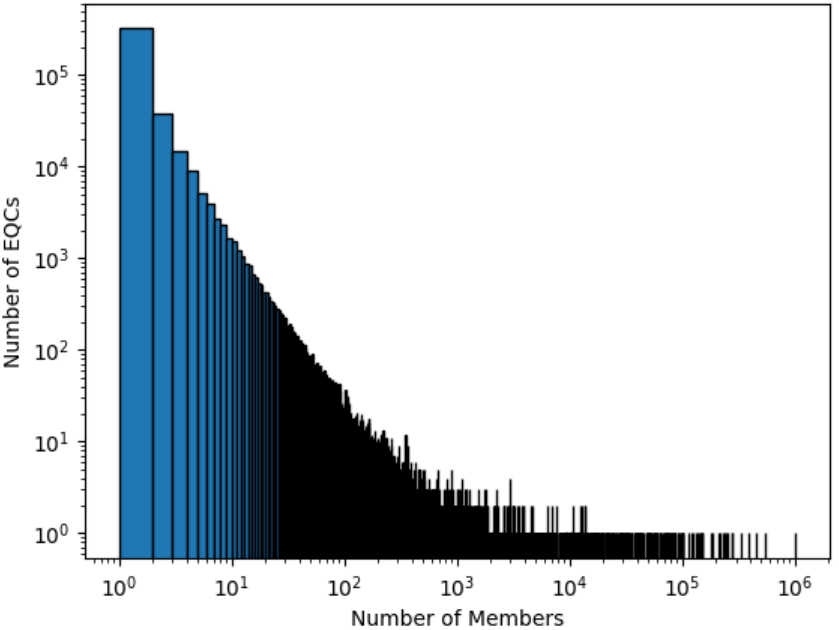}
         \caption{Frequency of number of members of an EQCs (in descending order) for the $2$-hop attribute collection.}
     \end{subfigure}  
     \begin{subfigure}[b]{0.3\textwidth}
         \centering
         \includegraphics[width=\textwidth]{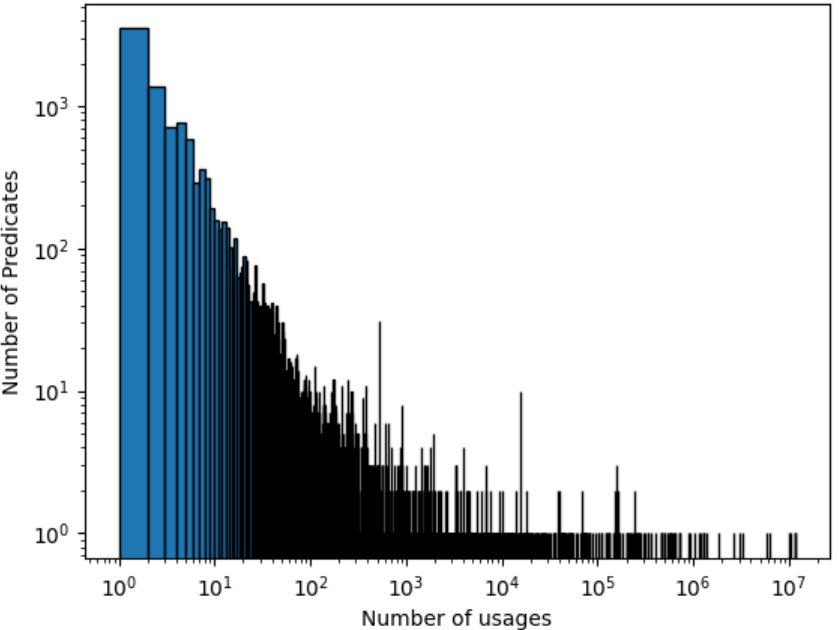}
         \caption{Frequency how often a predicate is used in the original graph (in descending order).}
     \end{subfigure}  \hfill
        \caption{Distributions measures on the summary of 2022-09-25}
        \label{fig:dist2022-09}
\end{figure}

Figure~\ref{fig:deviation} shows the average sizes of the EQCs as well as the average numbers of edges.
Example distributions are depicted in Figures~\ref{fig:dist2012-05} and~\ref{fig:dist2022-09} for one snapshot in 2012 and 2022, respectively.

\subsubsection{Binary Measures}
Binary measures are used to compare a current summary $S_{t+1}$ to the previous summary $S_{t}$. It is used to measure the change that occurred between $S_{t+1}$  and $S_{t}$. The following measure is calculated with the Jaccard coefficient to calculate the change of vertices:
\begin{equation}\delta^{(2)}_{JacV}(S_t,S_{t+1})= 1-\frac{|V^S_{t}\cap V^S_{t+1}|}{|V^S_{t}\cup V^S_{t+1}|}.\end{equation}

\begin{figure}[H]
     \centering
     \begin{subfigure}[b]{0.35\textwidth}
         \centering
         \includegraphics[width=\textwidth]{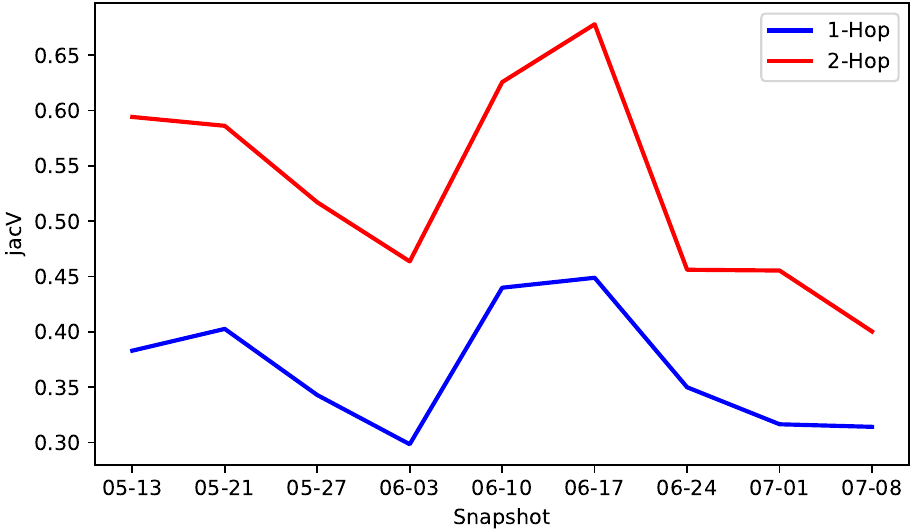}
         \caption{  Jaccard Coefficients in 2012.}
         \label{Jacc2012}
     \end{subfigure}  
     \begin{subfigure}[b]{0.35\textwidth}
         \centering
         \includegraphics[width=\textwidth]{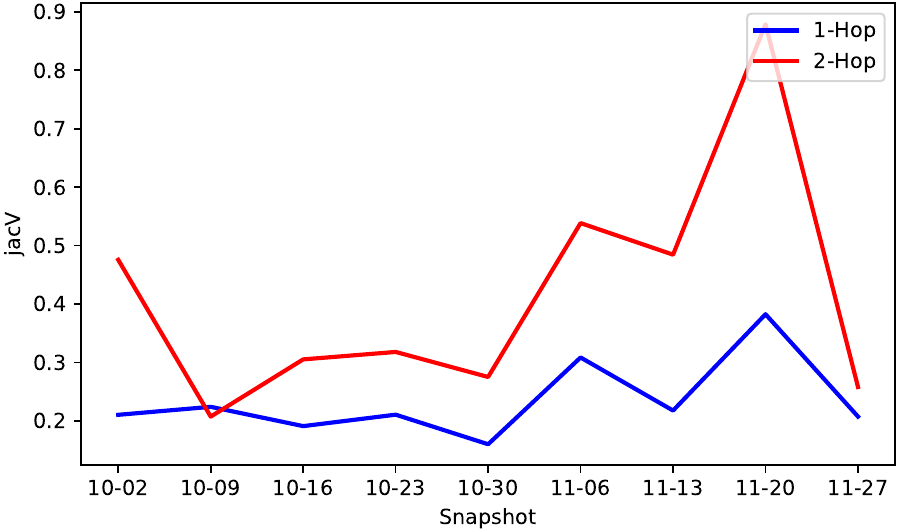}
         \caption{ Jaccard Coefficients in 2022.}
         \label{Jacc2022}
     \end{subfigure}  
     \begin{subfigure}[b]{0.35\textwidth}
         \centering
         \includegraphics[width=\textwidth]{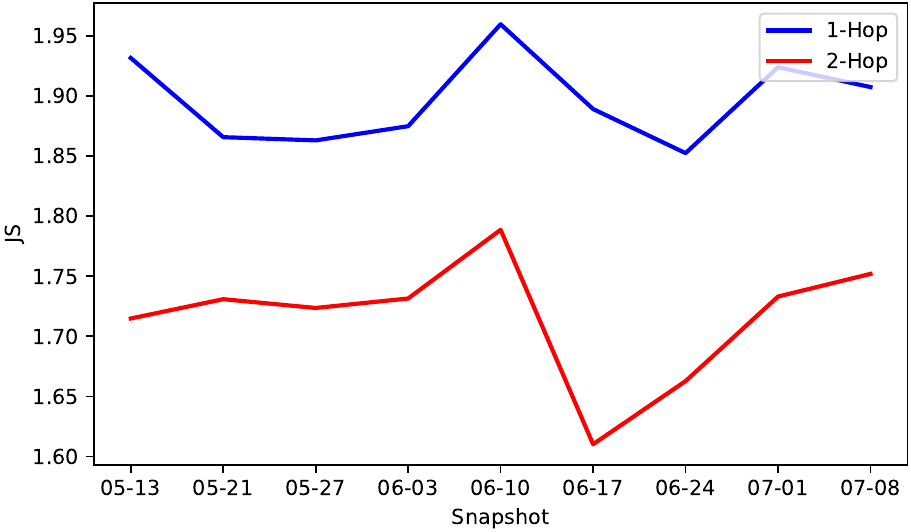}
         \caption{  Jensen-Shannon Divergences in 2012.}
         \label{JS2012}
     \end{subfigure}  
     \begin{subfigure}[b]{0.35\textwidth}
         \centering
         \includegraphics[width=\textwidth]{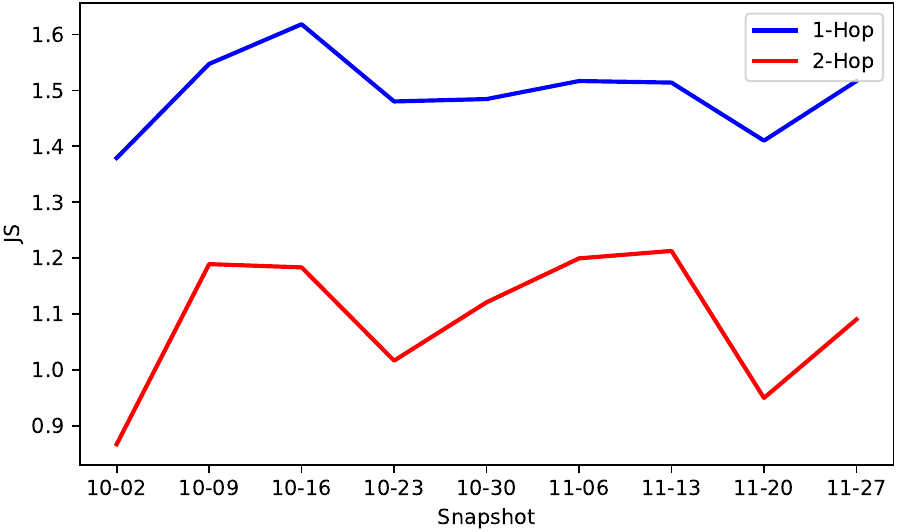}
         \caption{ Jensen-Shannon Divergences in 2022.}
         \label{JS2022}
     \end{subfigure}  \
        \caption{Jaccard Coefficients and Jensen-Shannon Divergences in the snapshots of 2012 and 2022.}
        \label{fig:binaryM}
\end{figure}

We also calculate the 
Jensen–Shannon Divergence of $S_{t+1}$ and $S_{t}$.  The Jensen–Shannon Divergence can be seen as the relative entropy
between two probability distributions and is calculated as 
\begin{equation}\delta^{(2)}_{JS}(A,B)=D(A,B)+D(B,A)\end{equation}
with \begin{equation}D(A,B)=\sum_{q\in V^B}P_{A}(q)\log_2 \frac{P_{A}(q)}{P_{B}(q)}\end{equation}
and 
$ P_{A}(q)= \frac{ext_t(q)}{|V_A|},$ where $V_A$ are the vertices in A. If $ext_t(q) =0$, then $P_{A}(eq) = 0$, and we consider the summand to be 0 as 
$\lim_{x \rightarrow 0} x \log (x) = 0.$
The changes in the binary measures are shown in Figure~\ref{fig:binaryM}.

\subsubsection{Meta-Measures}
\begin{figure}[H]
     \centering
     \begin{subfigure}[b]{0.24\textwidth}
         \centering
         \includegraphics[width=\textwidth]{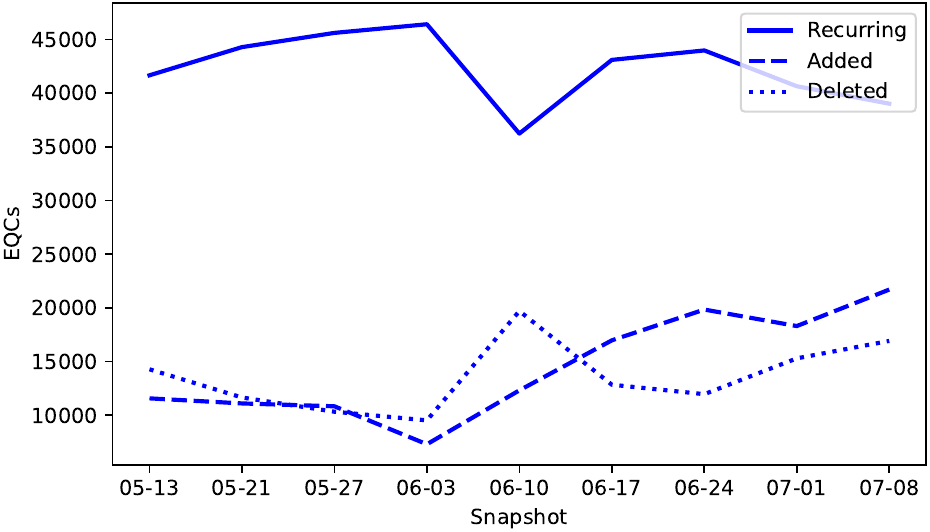}
         \caption{ Changes of the EQCs compared to the first one in 2012 ($1$-hop).}
         \label{ChangesF2012-1}
     \end{subfigure}  \hfill
     \begin{subfigure}[b]{0.24\textwidth}
         \centering
         \includegraphics[width=\textwidth]{images/1AC_changesPrev-1-cropped.pdf}
         \caption{  Changes of the EQCs compared to the previous one in 2012 ($1$-hop).}
         \label{ChangesP2012-1}
     \end{subfigure}  \hfill
      \begin{subfigure}[b]{0.24\textwidth}
         \centering
         \includegraphics[width=\textwidth]{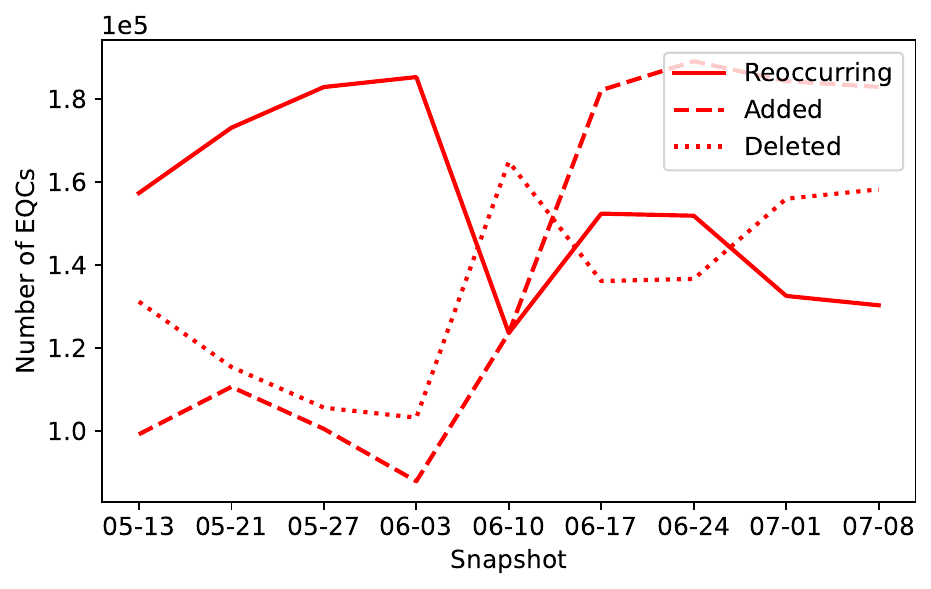}
         \caption{ Changes of the EQCs compared to the first one in 2012 ($2$-hop).}
         \label{ChangesF2012-2}
     \end{subfigure}  \hfill
     \begin{subfigure}[b]{0.24\textwidth}
         \centering
         \includegraphics[width=\textwidth]{images/2AC_changesPrev-1-cropped.pdf}
         \caption{ Changes of the EQCs compared to the previous one in 2012 ($2$-hop).}
         \label{ChangesP2012-2}
     \end{subfigure}  \hfill
     \begin{subfigure}[b]{0.24\textwidth}
         \centering
         \includegraphics[width=\textwidth]{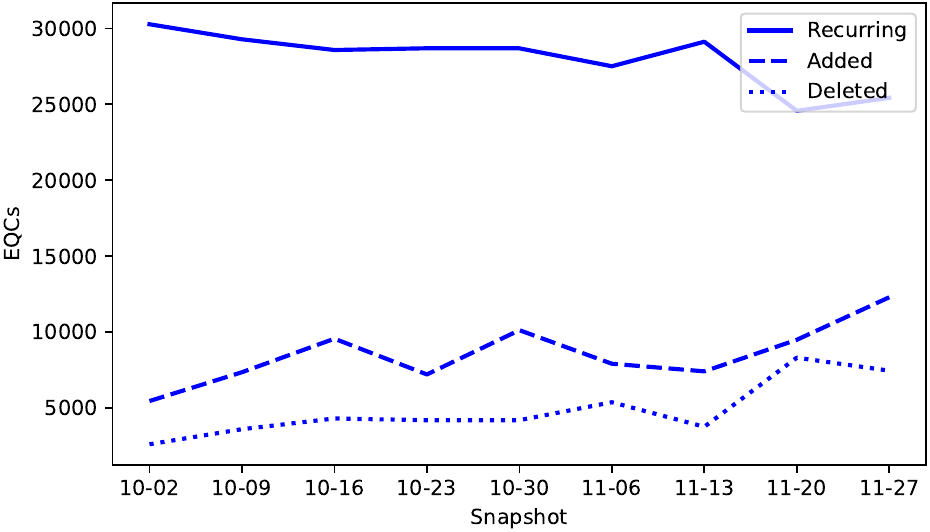}
          \caption{Changes of the EQCs compared to the first one in 2022 ($1$-hop).}
         \label{ChangesF2022-1}
     \end{subfigure}  \hfill
     \begin{subfigure}[b]{0.24\textwidth}
         \centering
         \includegraphics[width=\textwidth]{images/1AC_changesPrev-2-cropped.pdf}
         \caption{  Changes of the EQCs compared to the previous one in 2022 ($1$-hop).}
         \label{ChangesP2022-1}
     \end{subfigure}  \hfill
      \begin{subfigure}[b]{0.24\textwidth}
         \centering
         \includegraphics[width=\textwidth]{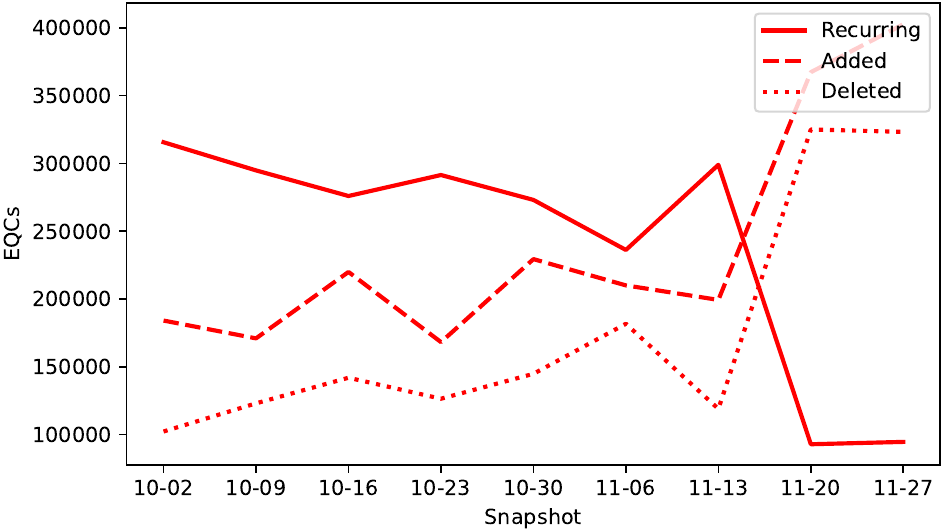}
           \caption{  Changes of the EQCs compared to the first one in 2022 ($2$-hop).}
         \label{ChangesF2022-2}
     \end{subfigure}  \hfill
     \begin{subfigure}[b]{0.24\textwidth}
         \centering
         \includegraphics[width=\textwidth]{images/2AC_changesPrev-2-cropped.pdf}
         \caption{  Changes of the EQCs compared to the previous one in 2022 ($2$-hop).}
         \label{ChangesP2022-2}
     \end{subfigure}  \hfill
     \begin{subfigure}[b]{0.24\textwidth}
         \centering
         \includegraphics[width=\textwidth]{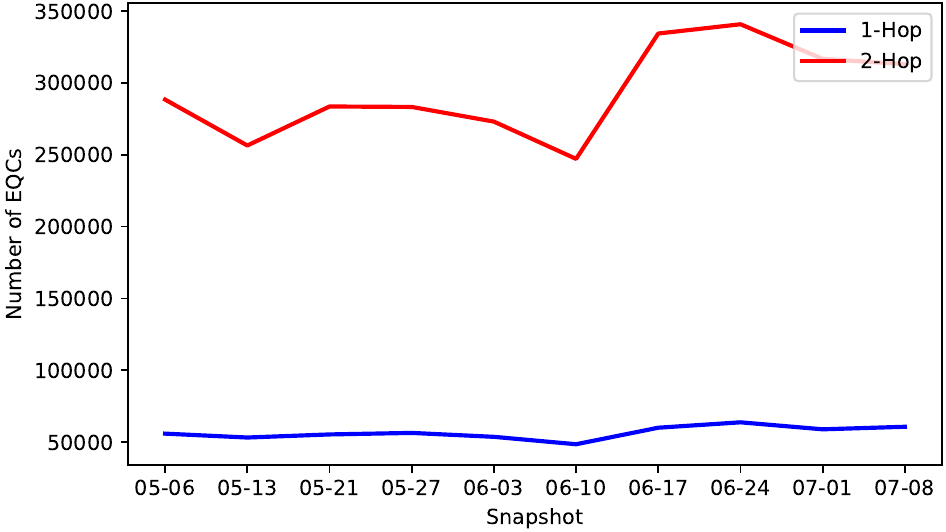}
         \caption{  Number of EQCs in each snapshot of 2012.}
         \label{EQCs2012}
     \end{subfigure}  \hfill
     \begin{subfigure}[b]{0.24\textwidth}
         \centering
         \includegraphics[width=\textwidth]{images/AC_EQCsAll-1-cropped.pdf}
          \caption{  Number of all seen EQCs till a snapshot in 2012.}
         \label{AllEQCs2012}
     \end{subfigure}  \hfill
     \begin{subfigure}[b]{0.24\textwidth}
         \centering
         \includegraphics[width=\textwidth]{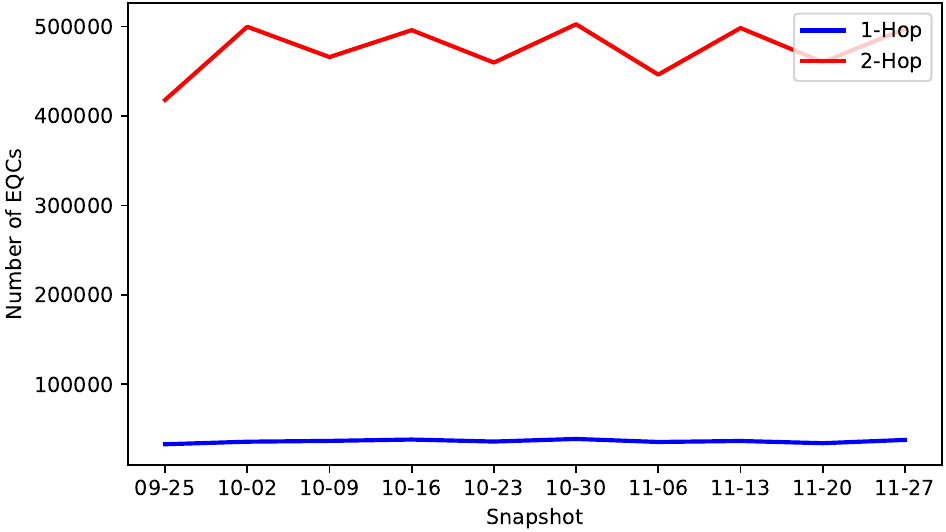}
         \caption{ Number of EQCs in each snapshot of 2022.}
         \label{EQCs2022}
     \end{subfigure}  \hfill
     \begin{subfigure}[b]{0.24\textwidth}
         \centering
         \includegraphics[width=\textwidth]{images/AC_EQCsAll-2-cropped.pdf}
          \caption{ Number of all seen EQCs till a snapshot in 2022.}
         \label{AllEQCs2022}
     \end{subfigure}  \hfill
        \caption{Measuring the changes in the EQCs per snapshot in 2012 and 2022.}
        \label{fig:changesAll}
\end{figure}

With meta-measures, we track how the summaries change over time, and in contrast to binary measures, it considers more than two summaries.
During our experiments, we track whether new EQCs are appearing compared to the first snapshot and the previous one. We also track whether EQCs are reappearing or disappearing and how many EQCs have been seen until a certain snapshot.
The meta-measures of our snapshots in 2012 and 2022 are depicted in Figure~\ref{fig:changesAll}.

\subsection{Impact of Layer Normalization}
\label{sec:impactofnormalization}

We investigate the impact of normalizing a GCN, we normalize the GCN by using the degrees of the vertices. 
The adapted activation $h^k_v$
of each vertex $v$ in layer $k$ is calculated
as follows:
\begin{equation}
    h^k_v = \sigma \left(\sum_{u\in V_s} \frac{1}{\sqrt{\hat{d}_v \hat{d}_u}} \tilde{A}_P[v,u](W^{k-1})^Th_u^{(k-1)} \right) \,,
\end{equation}
where $\hat{d}_v$ is the degree of $v$ and $\hat{d}_u$ is the degree of $u$. 

\begin{figure}[H]
\centering
\begin{subfigure}[b]{0.3\textwidth}
         \centering
\includegraphics[width=\textwidth]{images/accuracy_graphsaint_2-1-cropped.pdf}
         \caption{GCN ($2$-hop)}
         \label{CHeatmapGraphSAINT2E_2012}
         \end{subfigure} 
\begin{subfigure}[b]{0.3\textwidth}
         \centering
\includegraphics[width=\textwidth]{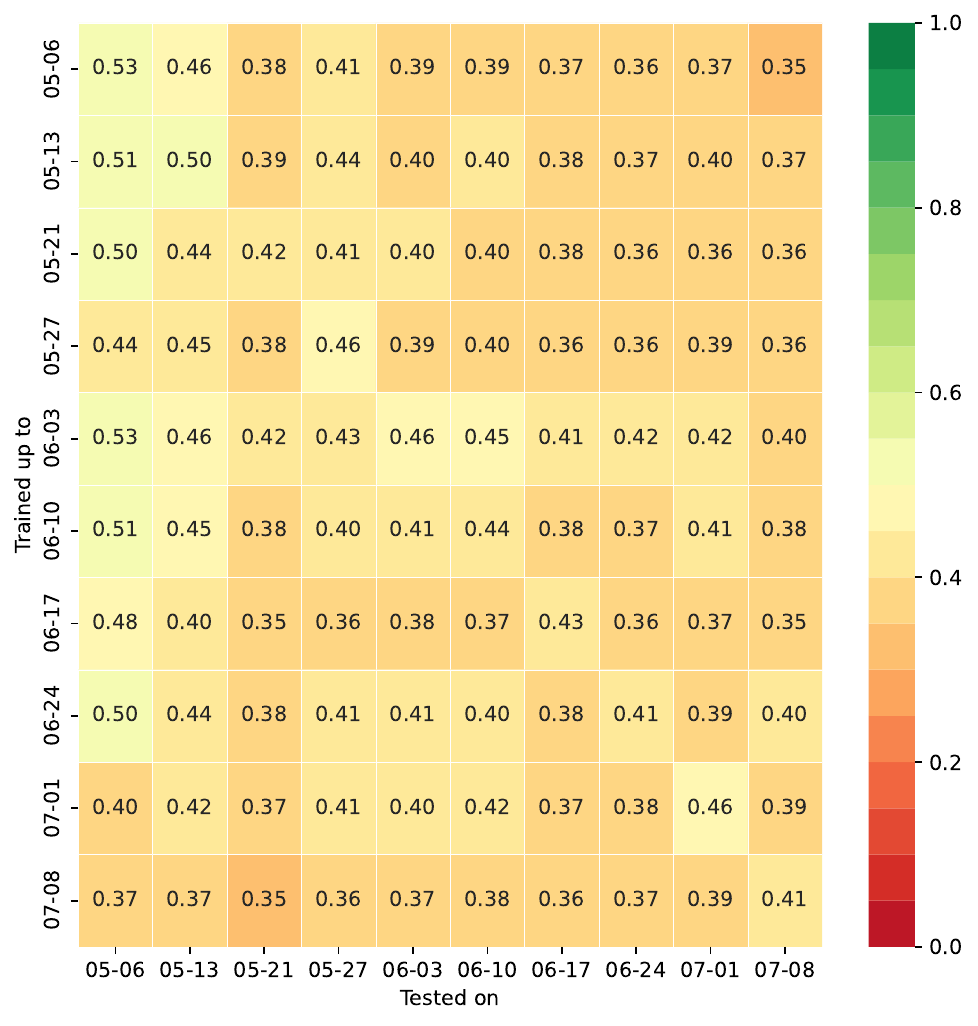}
         \caption{GCN ($2$-hop and normalized).}
         \label{CHeatmapGraphSAINT2N_2012}
         \end{subfigure}  \hfill
         \caption{Accuracies comparison (Normalization vs. no normalization)}
        \label{fig:heatsN}
\end{figure}

The results of the accuracy scores depicted in Figure~\ref{fig:heatsN} show that normalization does not make a big difference. 
This also is true of the forgetting scores, which are reported in Tables \ref{tab:llmeasuresN} and \ref{tab:forgettingN}. 

\begin{table}[H]\centering \small 
\begin{tabular}{|l|l|l|rrrrrr|} 
 \hline Year & Network & Type & ACC $\uparrow$ & BWT $\uparrow$ & FWT $\uparrow$ & $\Omega_\mathrm{base}$ $\uparrow$ & $\Omega_\mathrm{new}$ $\uparrow$ & $\Omega_\mathrm{all}$ $\uparrow$ \\ 
 \hline 2012 & GCN & $2$-hop & 0.376 & -0.087 & -0.043 & 0.824 & 0.445 & 0.672\\ 
       2012 & GCN & $2$-hop and normalized &  0.372 & -0.091 & -0.042 & 0.825 & 0.445 & 0.675\\ \hline \end{tabular}
 \caption{Lifelong Learning Measures (Normalization vs. no normalization).}\label{tab:llmeasuresN}
\end{table}

\begin{table}[H]\centering  \small 
\begin{tabular}{|l|l|l|rrrrrrrrr|} 
 \hline Year & Network & Type & $F_{2}$ & $F_{3}$ & $F_{4}$ & $F_{5}$ & $F_{6}$ & $F_{7}$ & $F_{8}$ & $F_{9}$ & $F_{10}$\\ 
 \hline 2012 & GCN & $2$-hop & 0.010 & 0.042 & 0.071 & 0.032 & 0.037 & 0.080 & 0.044 & 0.071 & 0.087\\ 
   2012 & GCN & $2$-hop and normalized & 0.019 & 0.047 & 0.064 & 0.017 & 0.047 & 0.081 & 0.049 & 0.065 & 0.093\\ 
 \hline \end{tabular}
 \caption{Forgetting values (Normalization vs. no normalization).}\label{tab:forgettingN}
\end{table}
    
\end{document}